\documentclass{article}

\PassOptionsToPackage{numbers,compress}{natbib}
    \usepackage[final]{neurips_2022}

\usepackage[utf8]{inputenc} 
\usepackage[T1]{fontenc}    
\usepackage[hidelinks]{hyperref}       
\usepackage{url}            
\usepackage{booktabs}       
\usepackage{amsfonts}       
\usepackage{nicefrac}       
\usepackage{microtype}      
\usepackage{xcolor}         
\usepackage{subfig}
\usepackage{graphicx}
\usepackage{amsmath}

\title{A Multi-Scale Deep Learning Framework for Projecting Weather Extremes}

\author{%
  Antoine Blanchard \\
  Verisk Analytics \\
  Boston, MA 02115 \\
  \texttt{ablanchard@verisk.com} \\
  \And
  Nishant Parashar \\
  Verisk Analytics \\
  Boston, MA 02115 \\
  \texttt{nparashar@verisk.com} \\
  \And
  Boyko Dodov \\
  Verisk Analytics \\
  Boston, MA 02115 \\
  \texttt{bdodov@verisk.com} \\
  \And
  Christian Lessig \\
  Otto-von-Guericke Universität\\
  Magdeburg, Germany \\
  \texttt{christian.lessig@ovgu.de} \\
  \And
  Themistoklis Sapsis \\
  Massachusetts Institute of Technology\\
  Cambridge, MA 02139 \\
  \texttt{sapsis@mit.edu} \\
}

\begin{document}

\maketitle

\begin{abstract}
Weather extremes are a major societal and economic hazard, claiming thousands of lives and causing billions of dollars in damage every year. Under climate change, their impact and intensity are expected to worsen significantly. Unfortunately, general circulation models (GCMs), which are currently the primary tool for climate projections, cannot characterize weather extremes accurately. 
To address this, we present a multi-resolution deep-learning framework that, firstly, corrects a GCM's biases by matching low-order and tail statistics of its output with observations at coarse scales; and secondly, increases the level of detail of the debiased GCM output by reconstructing the finer scales as a function of the coarse scales. We use the proposed framework to generate statistically realistic realizations of the climate over Western Europe from a simple GCM corrected using observational atmospheric reanalysis. We also discuss implications for  probabilistic risk assessment of natural disasters in a changing climate.
\end{abstract}

\section{Introduction}

The development of extreme weather under climate change is of growing concern for many stakeholders across the globe, including civil society, governments, and businesses~\citep{houser2015economic,field2012managing,sec2022disclosure}. The assessment of the risks associated with weather extremes is thereby highly critical.  Quite often, the most unlikely events in the tail of a hazard's probability distribution are also the most devastating.  Because of their unlikely nature, their statistics are difficult to predict, even more so on a regional scale~\citep{lucarini2016extremes}. In addition, the effects of climate change are expected to vary across weather perils, making exploration of catastrophic scenarios even more challenging.

Extreme weather risk is often estimated by generating ensembles of possible climate states using global circulation models (GCMs). Unfortunately, ensemble size and resolution are insufficient to characterize tail events properly.  Even at high resolution, GCMs suffer from statistical biases~\citep{wang2014global}
which can lead to the incorrect characterization of extreme-weather inducing processes (e.g., blocking events~\citep{davini2016northern}).  Moreover, high-resolution GCMs require vast computing resources.
Hence, to create large ensembles that provide a better statistical sampling, simple GCMs and coarse resolutions have to be employed. However, these simulations will be polluted with severe biases. 

To remedy these issues, one strategy is to leverage the wealth of observational data available about the Earth system and the power of machine learning~\citep{kashinath2021physics}. This can be pursued either in a purely data-based fashion~\citep{pathak2022fourcastnet}, or by augmenting the GCM dynamical equations with data-driven terms \cite{arcomano2022hybrid} or post-processing the GCM output to make it consistent with observations \citep{wan2021data,fulton2021towards}. The main challenges are thereby the strongly nonlinear energy scattering across scales as well the non-deterministic nature of the very fine scales \citep{sapsis2021statistics}.  These two issues often compromise the stability of machine-learning schemes \citep{brenowitz2020interpreting,yuval2020stable} and the physical consistency of the output \cite{beucler2021enforcing,chattopadhyay2021towards}.

We introduce a multi-resolution deep learning framework for accelerating the simulation of weather extremes. We combine a physics-based GCM run at coarse resolution with machine-learning models trained on observational data to reduce the biases and enhance the resolution of the GCM simulation.  The key ingredients are a) a compact, multi-scale representation of physical processes on the sphere which efficiently captures energy transfers across scales; b) novel statistical loss functions which place emphasis on extreme values and space--time coherency; and c) a divide-and-conquer training strategy which enables training of regional models at high spatial resolution.  The full-scale debiased simulation provides a complete view of risk at arbitrary resolution which decision makers can use to investigate near-present scenarios and assess their vulnerability to catastrophic weather events.

\section{Methods}

\begin{figure}[t]
  \centering
  \includegraphics[width=\textwidth, clip=true, trim=20 160 180 20]{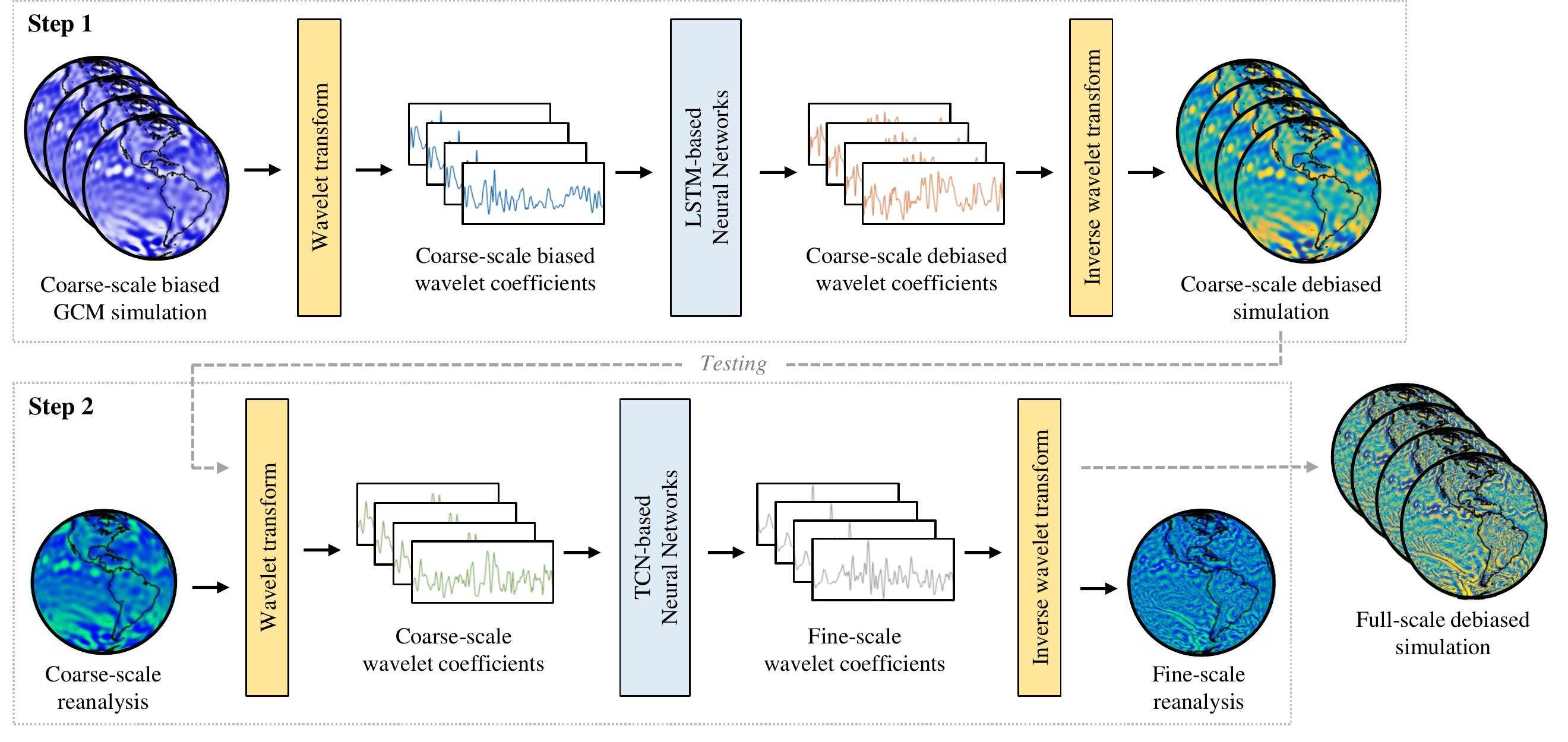}
  \caption{Overview of framework. Reanalysis data is used as benchmark as it provides the most comprehensive description of the observed climate.}
  \label{fig:fig1}
\end{figure}

We start from coarse-resolution runs of a simple but fast GCM polluted by strong biases. Based on these, we proceed in two steps; see Figure~\ref{fig:fig1}.  First, we correct the low-order and tail statistics of the GCM output at coarse scales to match the statistics of observations. Second, we enhance the resolution of the debiased GCM output by reconstructing the finer scales as a function of the coarse ones. In both steps, the ground truth is reanalysis data, a blend of observations and past forecasts which provides the most realistic picture of past atmospheric states that is available.

\paragraph{Decomposition of spatial scales}

\begin{figure}[t]
  \label{fig:fig2}
  \centering
  \subfloat[][\label{fig:fig2a}]{\includegraphics[height=0.4\textwidth, clip=true, trim=20 170 660 20]{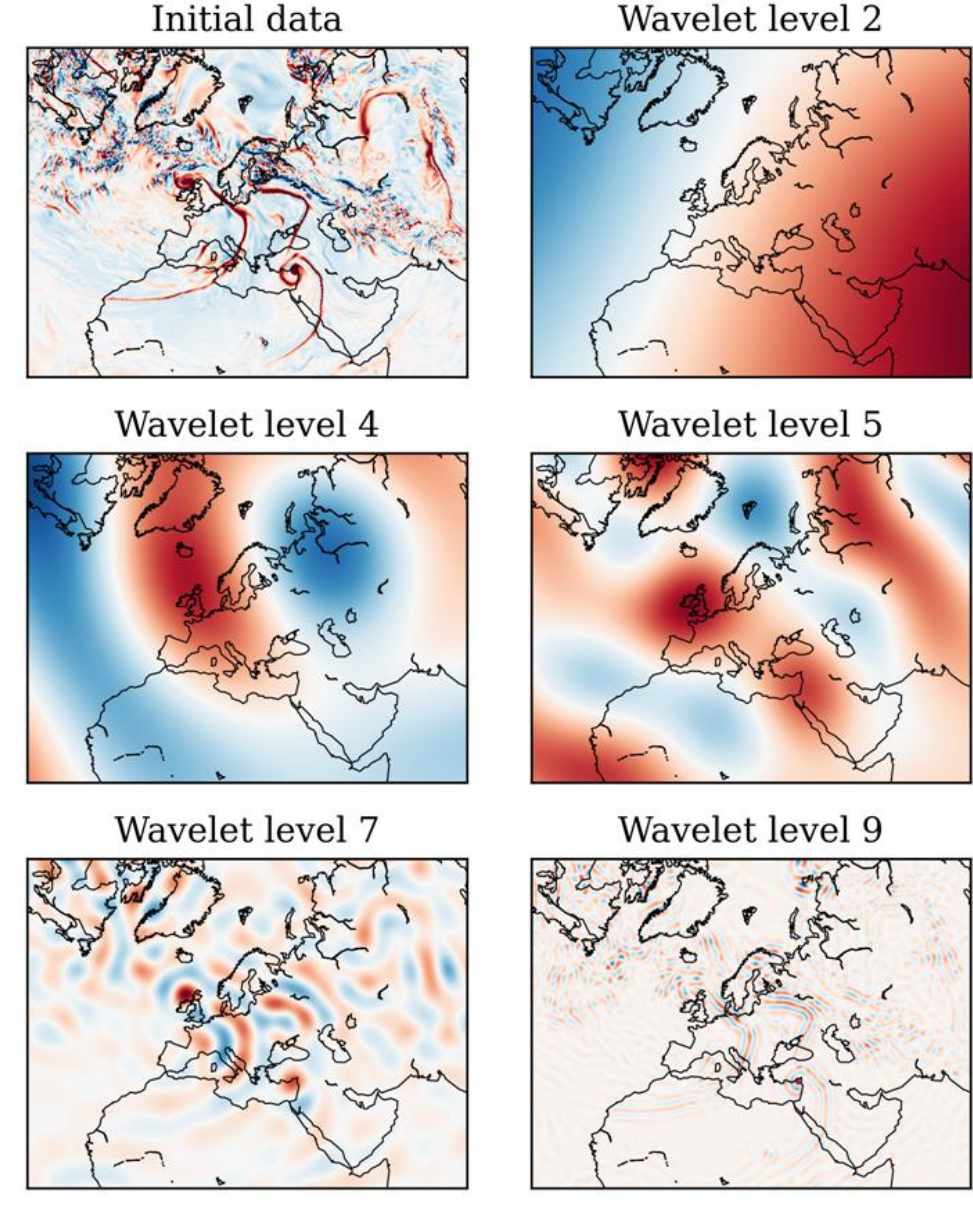}} \qquad
  \subfloat[][\label{fig:fig2b}]{\includegraphics[height=0.4\textwidth, clip=true, trim=10 170 480 10]{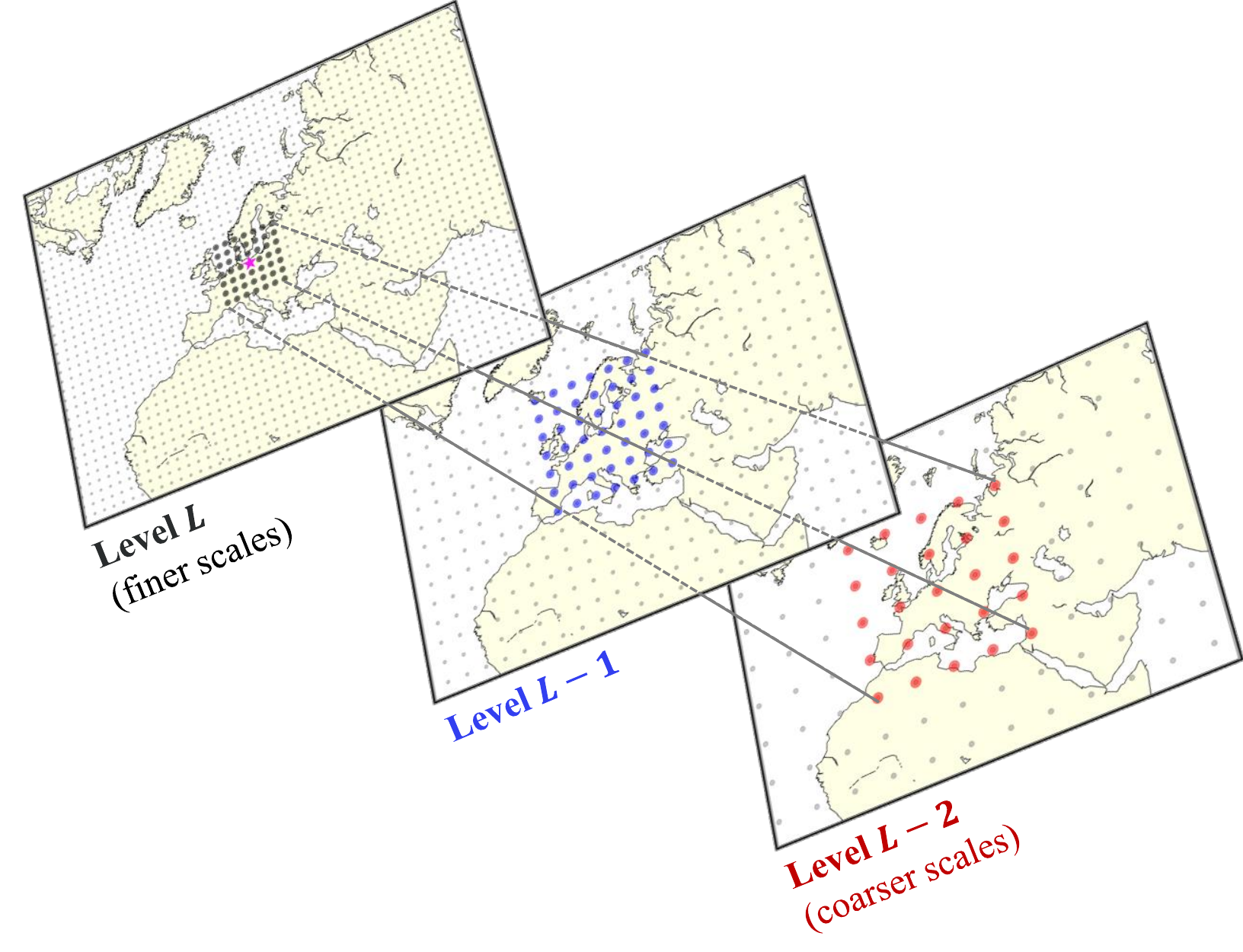}}
  \caption{(a) Wavelet decomposition for ERA5 reanalysis and (b) cone of influence for model input.}
\end{figure}

The dynamics of the atmosphere are characterized by a wide range of scales, from thousands of kilometers to a few meters. To delineate coarse- and fine-scale phenomena, we use a discrete spherical wavelet frame~\citep{da2020local} which provides a mathematically robust framework for scale separation and allows us to achieve a good balance between spatial and frequency localization. Each physical variable is decomposed on a set of wavelet levels of increasing granularity, with lower levels capturing large-scale spatial features (Figure \ref{fig:fig2a}). At each level, the locations of the wavelet centers are chosen as in \cite{mcewen2011novel} so as to guarantee the existence of a fast transform. This is important for manipulating very large climate ensembles at high spatial resolution. This also enables a highly efficient divide-and-conquer strategy whereby a multitude of small, local neural networks (one for each wavelet location) are trained concurrently, with communication between models being through a custom loss function (described below). More details are available in Section~\ref{sec:training}.

\paragraph{Step 1: Bias-correction of coarse GCM output}

The wavelet coefficients of the coarse, biased GCM output are mapped into a different set of wavelet coefficients whose statistics match those of coarse-scale observations; specifically, ERA5 reanalysis~\citep{hersbach2020era5}. The bias-correction step is formulated as a semi-supervised learning problem whereby reanalysis data is used as target for all GCM ensemble members. For each wavelet center, we employ an LSTM-based neural network whose input consists of time series for that center and its nearest neighbors. To capture energy transfers across scales---a landmark of turbulent flows---the input to the network also includes wavelet centers that belong to coarser wavelet levels.  This defines a cone of influence for the center under consideration (Figure \ref{fig:fig2b}). From this input, the model outputs a time series whose statistics are required to match those of reanalysis data. Once the models have been trained, the statistically-corrected time series are assembled and transformed back from wavelet to physical space to produce bias-corrected realizations of coarse-scale atmospheric dynamics.

\paragraph{Step 2: Reconstruction of small-scale features}

The ensemble obtained from Step 1 has the same resolution as the GCM on which it is based (typically, cells of size 250--500 km).  To achieve a higher resolution and be able to describe the local effects of extreme events, small-scale structures must be added.  This is done with a second machine-learning step whose task is to reconstruct the fine scales (i.e., those below the GCM resolution) as a function of the coarse scales (i.e., those resolved by the GCM).  The algorithm is trained on ERA5 reanalysis data in a supervised setting and tested on the debiased output from Step 1. Each wavelet center is assigned a TCN-based model that operates on a multi-scale cone of influence.  Wavelet levels can be trained in parallel, but this might lead to error accumulation across levels at testing time.  To combat this, we employ a sequential strategy akin to imitation learning \citep{venkatraman2015improving} whereby the input to each model is replaced with model predictions from previous levels.  This improves robustness of the predictions, especially for the coarser levels. 

\paragraph{Statistical loss functions} 

In Step 1, the models are trained using statistical loss functions; this is a key innovation of our framework. We consider two classes of statistical losses: quantile losses, which capture heavy tails and extreme values; and spectral losses, which capture cross-correlations between wavelet centers.  The former guarantee that the frequency and amplitude of extreme weather events in the output is consistent with reanalysis data.  The latter guarantee space--time coherency of the output with respect to wave propagation and transport phenomena. These losses are computed over several decades of weather data to ensure convergence of the statistics with respect to sample size (see Section~\ref{sec:losses}). Crucially, enforcing statistical consistency with reanalysis data still allows for variability of events in the output, especially the most extreme ones. In Step 2, statistical losses are used as regularizers, with the MSE loss (or a variant of it that emphasizes extreme values~\citep{rudy2021output}) being the main driver for the optimization.

\section{Results}
\label{sec:results}

\paragraph{Experimental protocol} 

To demonstrate the potential of the framework, we use SPEEDY \citep{molteni2003atmospheric,kucharski2006decadal} as our simplified GCM. The GCM runs are at the T47 resolution (about 300 km), thereby covering wavelet levels 1--5 in full. The target resolution for the downscaled output is about 80 km, corresponding to wavelet level 8. Our models are trained on vorticity and divergence close to the ocean surface ($\sigma$-level 0.95). For the debiasing step (Step 1), each model consists of a one-layer LSTM cell with hidden size 60 followed by a linear layer and a residual block.  Step 1 is trained on 5 GCM runs, each spanning 10 years; validation and testing are performed on two additional runs.  For the downscaling step (Step 2), each model consists of a TCN block with 4 channels of size 64 and kernel size 2. Step 2 is trained on 30 years of reanalysis data, and the remaining 10 years are used for validation and testing. Experiments are performed on Tesla V100 GPUs hosted on AWS. Training time on a single GPU is about 24 hours for Step 1 and about 60 hours for Step 2.  

\paragraph{Validation}

\begin{figure}[t]
  \centering
  \subfloat{\includegraphics[width=0.46\textwidth, clip=true, trim=0 0 100 0]{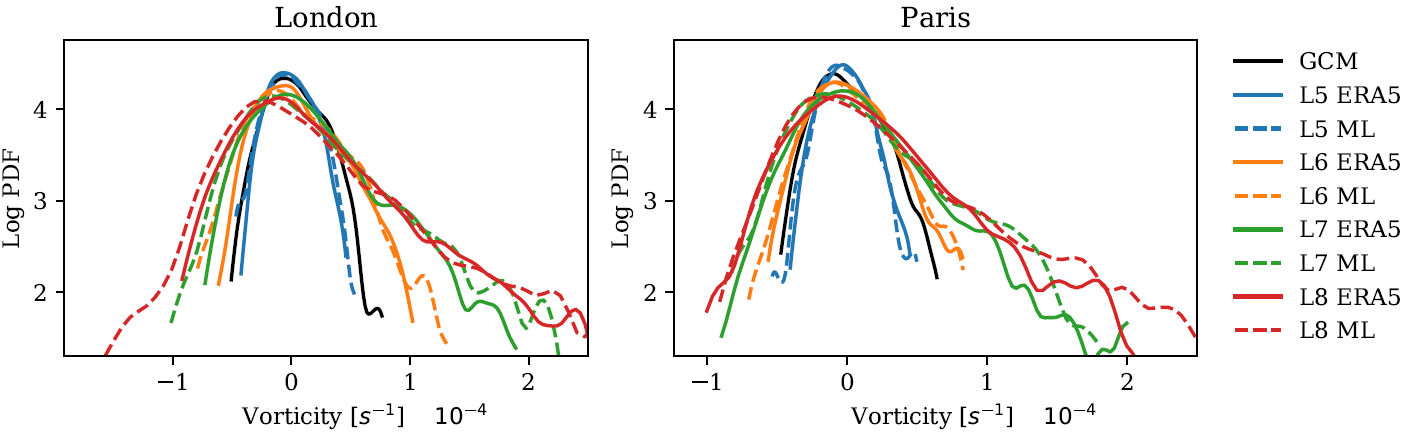}}
  \subfloat{\includegraphics[width=0.54\textwidth, clip=true, trim=0 0 0 0]{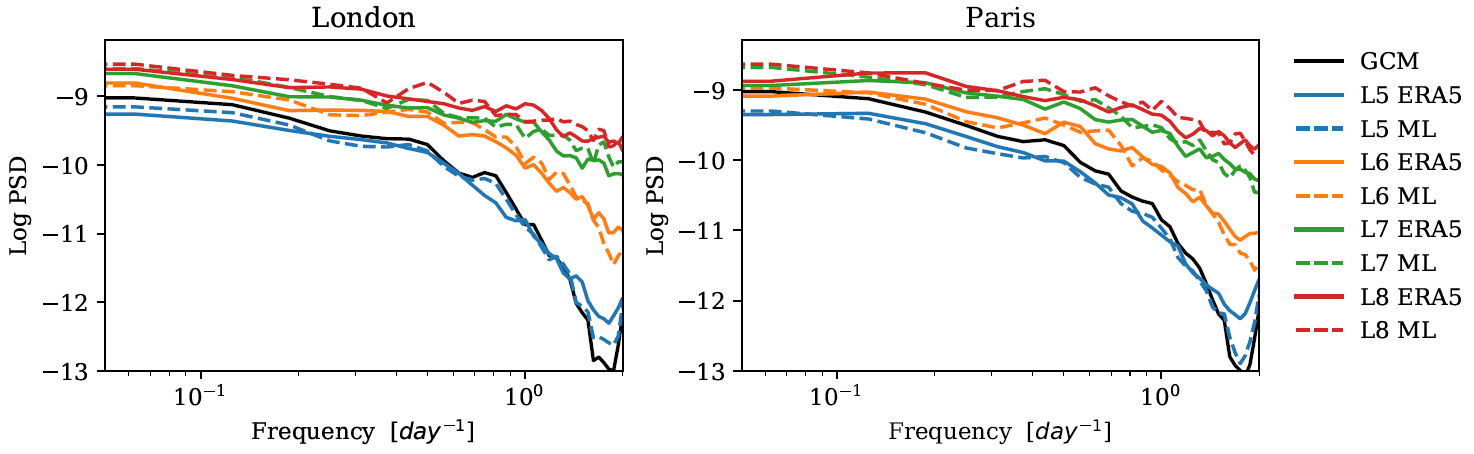}}\\
  \subfloat{\includegraphics[width=0.95\textwidth, clip=true, trim=20 360 340 20]{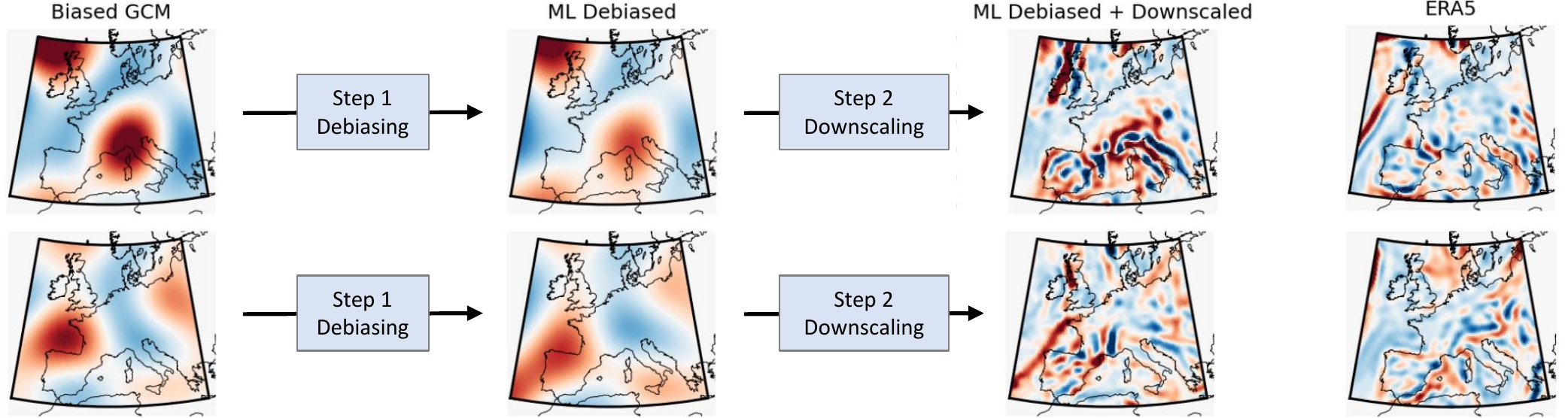}}
  \caption{Comparison of full-scale debiased simulation produced by the machine-learning algorithm to the SPEEDY GCM and reanalysis data. \textit{Top panel}: Temporal statistics (left: pdf; right: spectral density) of vorticity computed over a one-year period for Paris and London. \textit{Bottom panel}: 
  End-to-end illustration of the machine-learning pipeline tested on two snapshots of vorticity from the coarse-scale GCM; the ERA5 reanalysis snapshots on the far right are provided for \textit{qualitative} comparison only.}
  \label{fig:results}
\end{figure}

To assess performance, we transform the full-resolution wavelet time series that are produced by our algorithm back to physical space and consider a number of metrics defined at fixed locations across Western Europe.  Since our main goal is risk quantification, our success metrics are largely statistical in nature.  We also rely on visual inspection of movies by domain experts to evaluate realism of the machine-learning output. Figure~\ref{fig:results} shows that the debiasing models are able to correct the statistics of the GCM output at coarse scales. The fine-scale features added by the models from Step 2 bring the statistics of the full-resolution output even closer to the ERA5 ground truth. The machine-learning simulation looks remarkably realistic. Numerous fronts, which are not present in the coarse-scale GCM simulation, are clearly visible in the machine-learning output, emerging, propagating, and breaking as in the reanalysis.  Additional results are presented in Section~\ref{sec:moreresults}.

\section{Discussion}
\label{sec:discussion}

With the proposed framework, million-year simulations of weather extremes are becoming within reach, enabling more accurate quantification of catastrophic weather events. This will allow us to meet the growing demand from decision makers for global simulations that capture all types of dependencies across multiple regions and perils. These large ensembles can also be subsampled to create event simulations conditioned on a particular climate-change scenario (e.g., how would the combination of La Ni\~{n}a with a positive AMO index affect hurricane activity?).  This will considerably improve the estimation of the tail risks associated with weather extremes as they are affected by climate change.  From a technical standpoint, several aspects warrant further investigation.  In particular, it would be desirable to incorporate more physics into the machine-learning models, either in the form of soft constraints (e.g., a shallow-water PDE loss) or directly in the network architecture.

\begin{ack}
Funding in direct support of this work was provided by Verisk Analytics. CL acknowledges support from the Deutsche Forschungsgemeinschaft (DFG, German Research Foundation) – Project-ID 422037413 – TRR 287. The authors declare that they have no known competing financial interests or personal relationships that could have appeared to influence the submitted work.
\end{ack}


{
\small
\bibliography{refs}

\begin{thebibliography}{23}
\providecommand{\natexlab}[1]{#1}
\providecommand{\url}[1]{\texttt{#1}}
\expandafter\ifx\csname urlstyle\endcsname\relax
  \providecommand{\doi}[1]{doi: #1}\else
  \providecommand{\doi}{doi: \begingroup \urlstyle{rm}\Url}\fi

\bibitem[Houser et~al.(2015)Houser, Hsiang, Kopp, Larsen, Delgado, Jina,
  Mastrandrea, Mohan, Muir-Wood, Rasmussen, et~al.]{houser2015economic}
T.~Houser, S.~Hsiang, R.~Kopp, K.~Larsen, M.~Delgado, A.~Jina, M.~Mastrandrea,
  S.~Mohan, R.~Muir-Wood, D.~J. Rasmussen, et~al.
\newblock \emph{Economic risks of climate change: an American prospectus}.
\newblock Columbia University Press, 2015.

\bibitem[Field et~al.(2012)Field, Barros, Stocker, and Dahe]{field2012managing}
C.~B. Field, V.~Barros, T.~F. Stocker, and Q.~Dahe.
\newblock \emph{Managing the risks of extreme events and disasters to advance
  climate change adaptation: Special report of the Intergovernmental Panel on
  Climate Change}.
\newblock Cambridge University Press, 2012.

\bibitem[{The Securities and Exchange Commission}(March 21,
  2022)]{sec2022disclosure}
{The Securities and Exchange Commission}.
\newblock {SEC} proposes rules to enhance and standardize climate-related
  disclosures for investors.
\newblock \url{https://www.sec.gov/news/press-release/2022-46}, March 21, 2022.

\bibitem[Lucarini et~al.(2016)Lucarini, Faranda, de~Freitas, Holland, Kuna,
  Nicol, Todd, Vaienti, et~al.]{lucarini2016extremes}
V.~Lucarini, D.~Faranda, J.~M.~M. de~Freitas, M.~Holland, T.~Kuna, M.~Nicol,
  M.~Todd, S.~Vaienti, et~al.
\newblock \emph{Extremes and recurrence in dynamical systems}.
\newblock John Wiley \& Sons, 2016.

\bibitem[Wang et~al.(2014)Wang, Zhang, Lee, Wu, and Mechoso]{wang2014global}
C.~Wang, L.~Zhang, S.-K. Lee, L.~Wu, and C.~R. Mechoso.
\newblock A global perspective on {CMIP5} climate model biases.
\newblock \emph{Nature Climate Change}, 4\penalty0 (3):\penalty0 201--205,
  2014.

\bibitem[Davini and D’Andrea(2016)]{davini2016northern}
P.~Davini and F.~D’Andrea.
\newblock Northern hemisphere atmospheric blocking representation in global
  climate models: twenty years of improvements?
\newblock \emph{Journal of Climate}, 29\penalty0 (24):\penalty0 8823--8840,
  2016.

\bibitem[Kashinath et~al.(2021)Kashinath, Mustafa, Albert, Wu, Jiang,
  Esmaeilzadeh, Azizzadenesheli, Wang, Chattopadhyay, Singh,
  et~al.]{kashinath2021physics}
K.~Kashinath, M.~Mustafa, A.~Albert, J.~L. Wu, C.~Jiang, S.~Esmaeilzadeh,
  K.~Azizzadenesheli, R.~Wang, A.~Chattopadhyay, A.~Singh, et~al.
\newblock Physics-informed machine learning: case studies for weather and
  climate modelling.
\newblock \emph{Philosophical Transactions of the Royal Society A},
  379\penalty0 (2194):\penalty0 20200093, 2021.

\bibitem[Pathak et~al.(2022)Pathak, Subramanian, Harrington, Raja,
  Chattopadhyay, Mardani, Kurth, Hall, Li, Azizzadenesheli,
  et~al.]{pathak2022fourcastnet}
J.~Pathak, S.~Subramanian, P.~Harrington, S.~Raja, A.~Chattopadhyay,
  M.~Mardani, T.~Kurth, D.~Hall, Z.~Li, K.~Azizzadenesheli, et~al.
\newblock {FourCastNet: A global data-driven high-resolution weather model
  using adaptive Fourier neural operators}.
\newblock \emph{arXiv preprint arXiv:2202.11214}, 2022.

\bibitem[Arcomano et~al.(2022)Arcomano, Szunyogh, Wikner, Pathak, Hunt, and
  Ott]{arcomano2022hybrid}
T.~Arcomano, I.~Szunyogh, A.~Wikner, J.~Pathak, B.~R. Hunt, and E.~Ott.
\newblock A hybrid approach to atmospheric modeling that combines machine
  learning with a physics-based numerical model.
\newblock \emph{Journal of Advances in Modeling Earth Systems}, 14\penalty0
  (3):\penalty0 e2021MS002712, 2022.

\bibitem[Wan et~al.(2021)Wan, Dodov, Lessig, Dijkstra, and Sapsis]{wan2021data}
Z.~Y. Wan, B.~Dodov, C.~Lessig, H.~Dijkstra, and T.~P. Sapsis.
\newblock A data-driven framework for the stochastic reconstruction of
  small-scale features with application to climate data sets.
\newblock \emph{Journal of Computational Physics}, 442:\penalty0 110484, 2021.

\bibitem[Fulton et~al.(2021)Fulton, Clarke, Hegerl, and
  Otto]{fulton2021towards}
J.~Fulton, B.~Clarke, G.~Hegerl, and F.~Otto.
\newblock Towards debiasing climate simulations using image-to-image
  translation networks.
\newblock In \emph{NeurIPS 2021 Workshop: Tackling Climate Change with Machine
  Learning}, 2021.

\bibitem[Sapsis(2021)]{sapsis2021statistics}
T.~P. Sapsis.
\newblock Statistics of extreme events in fluid flows and waves.
\newblock \emph{Annual Reviews of Fluid Mechanics}, 58:\penalty0 85--111, 2021.

\bibitem[Brenowitz et~al.(2020)Brenowitz, Beucler, Pritchard, and
  Bretherton]{brenowitz2020interpreting}
N.~D. Brenowitz, T.~Beucler, M.~Pritchard, and C.~S. Bretherton.
\newblock Interpreting and stabilizing machine-learning parametrizations of
  convection.
\newblock \emph{Journal of the Atmospheric Sciences}, 77\penalty0
  (12):\penalty0 4357--4375, 2020.

\bibitem[Yuval and O’Gorman(2020)]{yuval2020stable}
J.~Yuval and P.~A. O’Gorman.
\newblock Stable machine-learning parameterization of subgrid processes for
  climate modeling at a range of resolutions.
\newblock \emph{Nature Communications}, 11\penalty0 (1):\penalty0 1--10, 2020.

\bibitem[Beucler et~al.(2021)Beucler, Pritchard, Rasp, Ott, Baldi, and
  Gentine]{beucler2021enforcing}
T.~Beucler, M.~Pritchard, S.~Rasp, J.~Ott, P.~Baldi, and P.~Gentine.
\newblock Enforcing analytic constraints in neural networks emulating physical
  systems.
\newblock \emph{Physical Review Letters}, 126\penalty0 (9):\penalty0 098302,
  2021.

\bibitem[Chattopadhyay et~al.(2021)Chattopadhyay, Mustafa, Hassanzadeh, Bach,
  and Kashinath]{chattopadhyay2021towards}
A.~Chattopadhyay, M.~Mustafa, P.~Hassanzadeh, E.~Bach, and K.~Kashinath.
\newblock Towards physically consistent data-driven weather forecasting:
  Integrating data assimilation with equivariance-preserving deep spatial
  transformers.
\newblock \emph{arXiv preprint arXiv:2103.09360}, 2021.

\bibitem[da~Silva et~al.(2020)da~Silva, Lessig, Dodov, Dijkstra, and
  Sapsis]{da2020local}
C.~C. da~Silva, C.~Lessig, B.~Dodov, H.~Dijkstra, and T.~Sapsis.
\newblock A local spectral exterior calculus for the sphere and application to
  the shallow water equations.
\newblock \emph{arXiv preprint arXiv:2005.03598}, 2020.

\bibitem[McEwen and Wiaux(2011)]{mcewen2011novel}
J.~D. McEwen and Y.~Wiaux.
\newblock A novel sampling theorem on the sphere.
\newblock \emph{IEEE Transactions on Signal Processing}, 59\penalty0
  (12):\penalty0 5876--5887, 2011.

\bibitem[Hersbach et~al.(2020)Hersbach, Bell, Berrisford, Hirahara,
  Hor{\'a}nyi, Mu{\~n}oz-Sabater, Nicolas, Peubey, Radu, Schepers,
  et~al.]{hersbach2020era5}
H.~Hersbach, B.~Bell, P.~Berrisford, S.~Hirahara, A.~Hor{\'a}nyi,
  J.~Mu{\~n}oz-Sabater, J.~Nicolas, C.~Peubey, R.~Radu, D.~Schepers, et~al.
\newblock The {ERA5} global reanalysis.
\newblock \emph{Quarterly Journal of the Royal Meteorological Society},
  146\penalty0 (730):\penalty0 1999--2049, 2020.

\bibitem[Venkatraman et~al.(2015)Venkatraman, Hebert, and
  Bagnell]{venkatraman2015improving}
A.~Venkatraman, M.~Hebert, and J.~A. Bagnell.
\newblock Improving multi-step prediction of learned time series models.
\newblock In \emph{Twenty-Ninth AAAI Conference on Artificial Intelligence},
  2015.

\bibitem[Rudy and Sapsis(2021)]{rudy2021output}
S.~Rudy and T.~Sapsis.
\newblock Output-weighted and relative entropy loss functions for deep learning
  precursors of extreme events.
\newblock \emph{arXiv preprint arXiv:2112.00825}, 2021.

\bibitem[Molteni(2003)]{molteni2003atmospheric}
F.~Molteni.
\newblock Atmospheric simulations using a {GCM} with simplified physical
  parametrizations. {I}: Model climatology and variability in multi-decadal
  experiments.
\newblock \emph{Climate Dynamics}, 20\penalty0 (2):\penalty0 175--191, 2003.

\bibitem[Kucharski et~al.(2006)Kucharski, Molteni, and
  Bracco]{kucharski2006decadal}
F.~Kucharski, F.~Molteni, and A.~Bracco.
\newblock Decadal interactions between the western tropical {P}acific and the
  {N}orth {A}tlantic {O}scillation.
\newblock \emph{Climate Dynamics}, 26\penalty0 (1):\penalty0 79--91, 2006.

\end{thebibliography}
}

\appendix

\section{Appendix}
\label{appendix}

\subsection{Training strategy}
\label{sec:training}

The spherical wavelet decomposition provides a multi-scale representation of physical variables on the sphere. Beside scale separation, it is also used to reduce the dimensionality of the input. At full ERA5 resolution, the spatial field at each time step is represented on a 721-by-1440 grid.  In wavelet space, the same spatial field can be represented almost losslessly using 9 wavelet levels and fewer than 700,000 wavelet coefficients, a nearly 30\% reduction compared to the physical grid.

Despite this compression, the size of the datasets on which the machine-learning algorithm is trained can be large due to the temporal resolution (3-hourly) and extent (several years) of the time series for each wavelet coefficient.  This bottleneck is exacerbated by the fact that the GCM dataset is an ensemble of realizations. This makes it impossible to fit the entire dataset into the memory of a single GPU. Multi-GPU training would be one solution but would also ultimately become impractical as more physical variables are added to the training. Another alternative would be to split the time series into small (e.g., monthly) chunks for mini-batching. However, this is incompatible with the statistical loss functions used in Step 1 as the latter require thousands of samples.

\begin{figure}[t]
  \centering
  \includegraphics[width=0.9\textwidth, clip=true, trim=20 220 420 20]{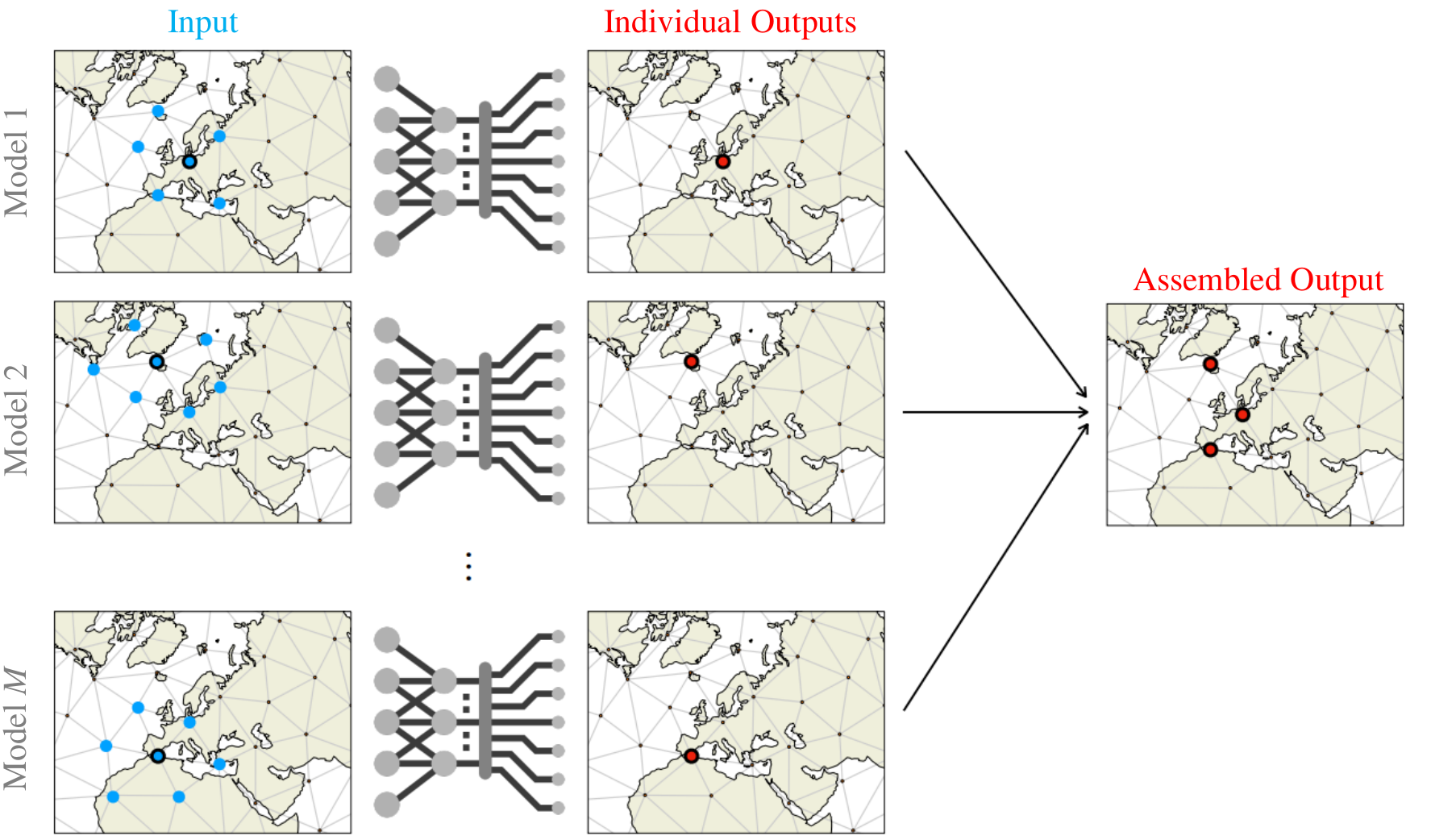}
  \caption{Overview of the divide-and-conquer strategy used for training.  In addition to alleviating the memory load, this approach allows the training of regional neural network models at high spatial resolution.}
  \label{fig:figA1}
\end{figure}

\begin{figure}[t]
  \centering
  \includegraphics[width=0.6\textwidth, clip=true, trim=20 240 500 20]{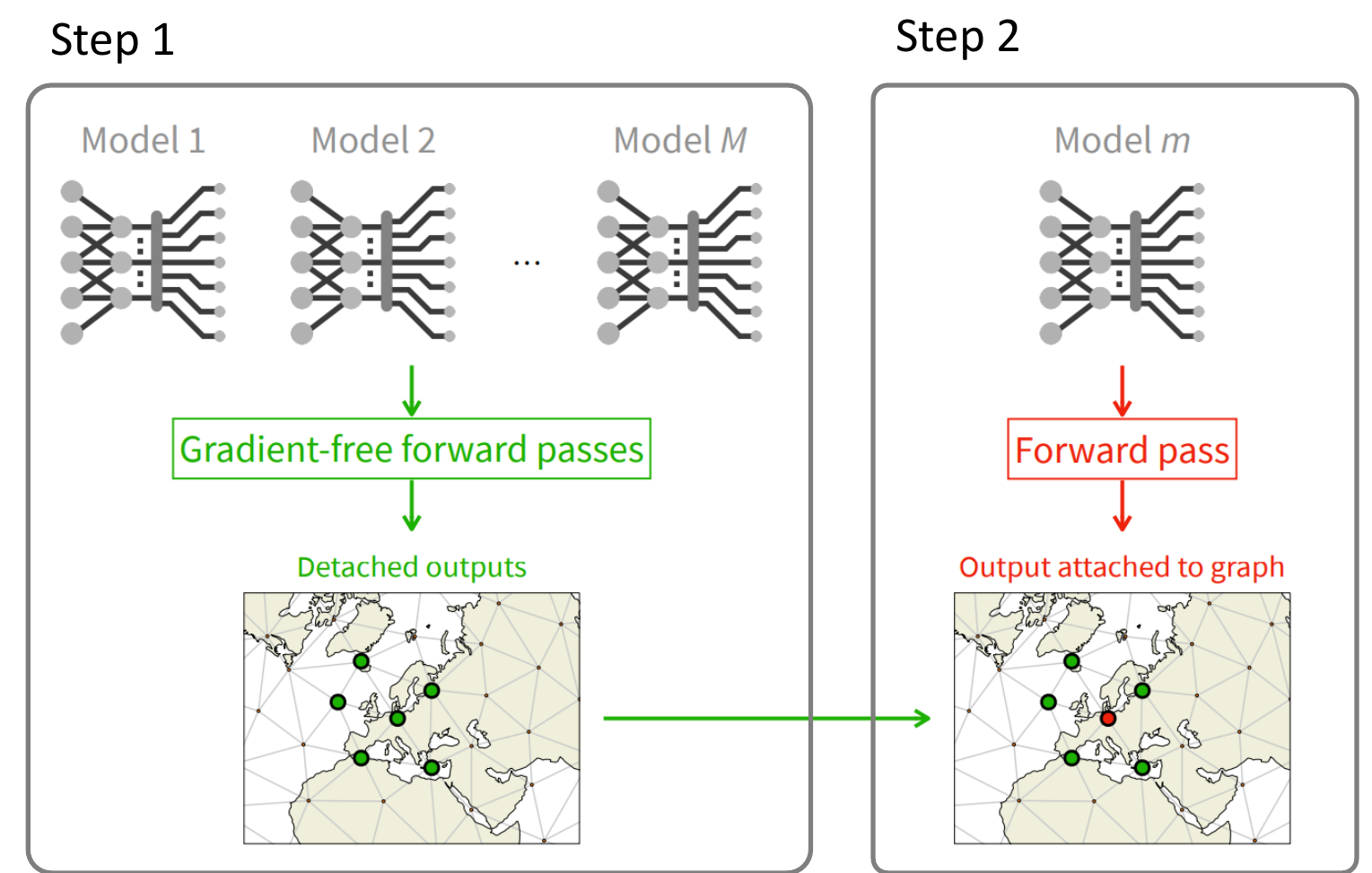}
  \caption{Sketch of one optimization step with forward passes applied sequentially without and with gradient computations.}
  \label{fig:figA2}
\end{figure}

To circumvent the memory issue, we deploy a divide-and-conquer strategy (Figure \ref{fig:figA1}) whereby a multitude of small, local models (one for each wavelet center) are trained concurrently, with communication between models being through the loss function. This strategy is computationally more efficient and less memory-intensive than one in which all wavelet centers are handled by one large model. To minimize the memory footprint, each optimization step consists of one gradient-free forward pass through all the models, combined with a series of gradient-equipped forward passes executed in sequence for each individual model (Figure \ref{fig:figA2}).  In this way, we never store more than one computational graph in memory at any given time. Conceptually, this strategy is similar to locally freezing certain components of one large model.

\subsection{Statistical loss functions}
\label{sec:losses}

\paragraph{Computational considerations}

We use statistical loss functions that ensure that the temporal statistics of the machine learning output are consistent with the observation-based reanalysis. Consistency is enforced in wavelet space, i.e., between the time series of wavelet coefficients.  The statistical losses operate on a single wavelet center $y(t)$, where we have a quantile loss, as well on neighborhoods of wavelet centers $\mathbf{y}(t)$, where we use a cross-spectrum loss. For a given statistical property $S$, our loss functions $\mathcal{L}$ take the form
\begin{equation}
\mathcal{L}(y_\textrm{out}, y_\textrm{obs}) = \textrm{MSE}(S(y_\textrm{out}), S(y_\textrm{obs})),
\end{equation}
with the subscripts ``out'' and ``obs'' denoting the machine-learning output and the observed value from reanalysis, respectively. To capture the temporal statistics accurately, the time series used in the loss function must contain a sufficiently large number of samples.  In Step 1, the times series used to compute the losses extend over a 10-year interval, which is computationally made possible by the divide-and-conquer strategy described above that alleviates memory issues.  In Step 2, the statistical losses are mainly used as regularizers and shorter (1-year) time series can be used. 

\paragraph{Quantile loss}

The quantile loss matches the quantiles of an individual machine-learned time series with its reanalysis counterpart.  It is defined as 
\begin{equation}
\mathcal{L}(y_\textrm{out}, y_\textrm{obs}) = \textrm{MSE}(\mathbf{q}(y_\textrm{out}),\mathbf{q}(y_\textrm{obs})),
\label{eq:qloss}
\end{equation}
where $\mathbf{q}$ is a vector containing the q-th quantiles of $y(t)$.  If $\mathbf{q}$ is specified as a vector of evenly spaced numbers between 0 and 1, then the quantile loss is equivalent to matching the marginal distributions of the individual time series. In practice we have found that no more than a dozen quantiles are required to properly capture the distributions.  Importantly, since we are interested in extreme events, we place special emphasis on the distribution tails  by using a non-uniform spacing of the quantiles; for instance,
\begin{equation}
\setcounter{MaxMatrixCols}{20}
\mathbf{q} = \begin{bmatrix} 0 & 0.001 & 0.01 & 0.1 & 0.3 & 0.5 & 0.7 & 0.9 & 0.99 & 0.999 & 1 \end{bmatrix} .
\end{equation}
We also tried a version of (\ref{eq:qloss}) similar to the cross-entropy loss but the results were inferior compared to (\ref{eq:qloss}).

\paragraph{Cross-spectrum loss}

Temporal dependencies between two real time series $x(t)$ and $y(t)$ is measured by the cross-correlation, defined as the temporal convolution between $x$ and $y$:
\begin{equation}
    a_{x,y}(\tau)= \int_{-\infty}^{\infty} x(t)y(t+\tau)\,\mathrm{d}t.
\end{equation}
Since the Fourier transform diagonalizes the convolution operator, one can equivalently compute the cross-spectrum $\Gamma_{x,y}$ between two signals $x$ and $y$:
\begin{equation}
    \Gamma_{x,y}(\omega)=\mathcal{F}[a_{x,y}](\omega),
\end{equation}
where $\mathcal{F}$ denotes the Fourier transform and $\omega$ a temporal frequency. The cross-spectrum is a complex-valued quantity and therefore can be written in polar form as
\begin{equation}
    \Gamma_{x,y}(\omega) = A_{x,y}(\omega) \exp(\mathrm{j}\Phi_{x,y}(\omega)),
\end{equation}
with $A$ denotes the real-valued cross-spectrum amplitude and $\Phi$ the phase.  The former quantifies how energy is distributed among frequencies and the latter contains information about how these frequencies align with each other in time.  Hence, the cross-spectrum can be used to capture propagation of wave and energy between neighboring wavelet centers.

We therefore define the cross-spectrum loss between a particular center and its neighbors as 
\begin{equation}
\mathcal{L}(y_\textrm{out}, y_\textrm{obs}) = \textrm{MSE}(A_{y_\textrm{out}, \mathbf{y}_\textrm{out}}(\omega), A_{y_\textrm{obs}, \mathbf{y}_\textrm{obs}}(\omega)) + \textrm{MSE}(\Phi_{y_\textrm{out},\mathbf{y}_\textrm{out}}(\omega), \Phi_{y_\textrm{obs},\mathbf{y}_\textrm{obs}}(\omega)).
\end{equation}
In practice, we found that penalizing real and imaginary parts to be more robust during training, and we therefore prefer
\begin{align}
\mathcal{L}(y_\textrm{out}, y_\textrm{obs}) &= \textrm{MSE}(\mathrm{Re}[\Gamma_{y_\textrm{out},\mathbf{y}_\textrm{out}}(\omega)],  \mathrm{Re}[\Gamma_{y_\textrm{obs},\mathbf{y}_\textrm{obs}}(\omega)])    \nonumber  \\
 & + \textrm{MSE}(\mathrm{Im}[\Gamma_{y_\textrm{out},\mathbf{y}_\textrm{out}}(\omega)], \mathrm{Im}[\Gamma_{y_\textrm{obs},\mathbf{y}_\textrm{obs}}(\omega)]).
\end{align}
Additional emphasis can be placed on the high frequencies by using a log-transform of $\Gamma$ in the loss function.  Note that if $y$ is appended to the vector of neighbors $\mathbf{y}$, then the cross-spectrum loss also captures the auto-correlation between the wavelet center $y$ and itself.  

\subsection{Additional results}
\label{sec:moreresults}

In this appendix we present additional results that consider the two steps of our algorithm separately. 

\begin{figure}[t]
  \centering
  \subfloat{\includegraphics[width=0.48\textwidth]{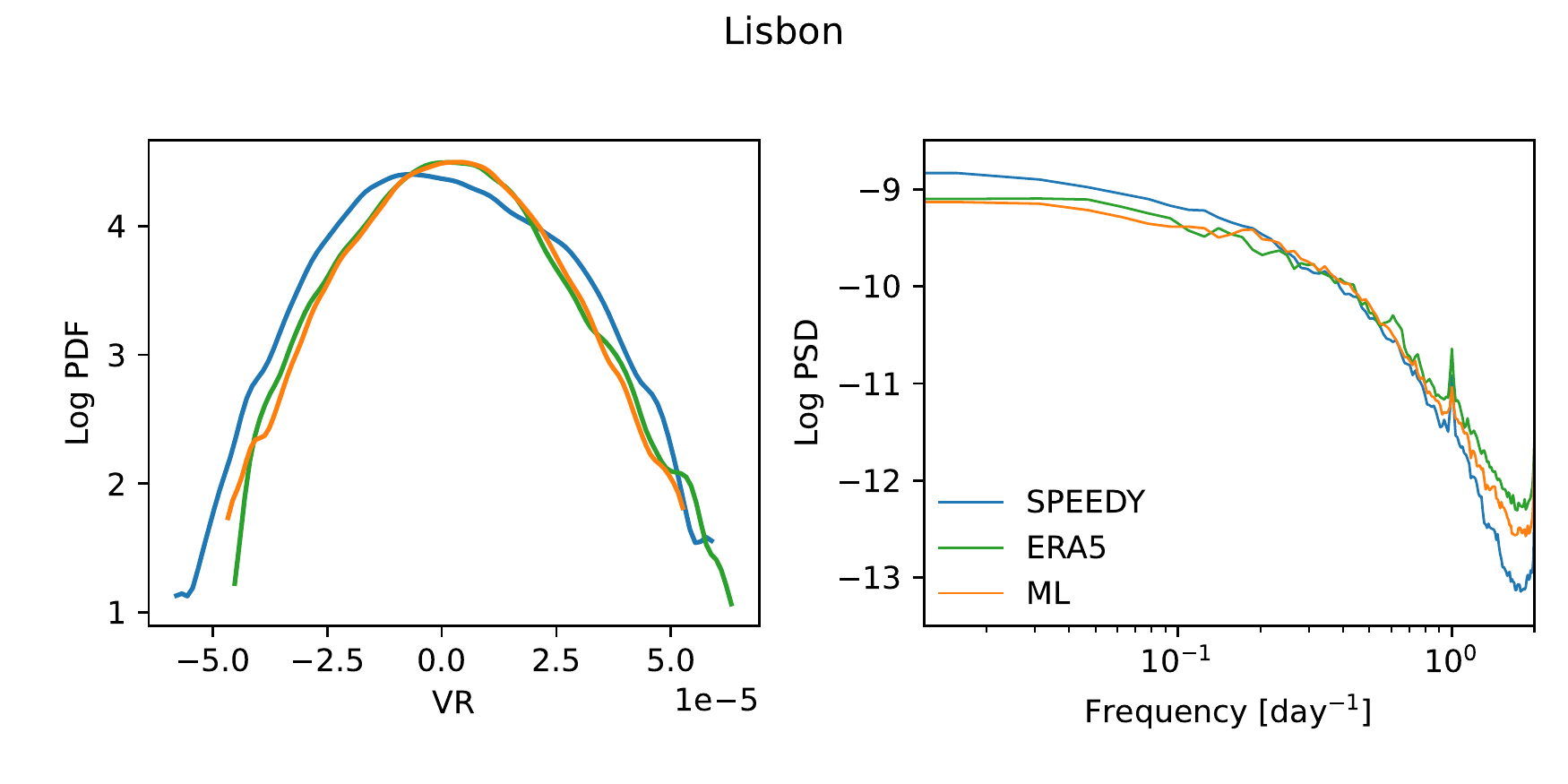}}
  \subfloat{\includegraphics[width=0.48\textwidth]{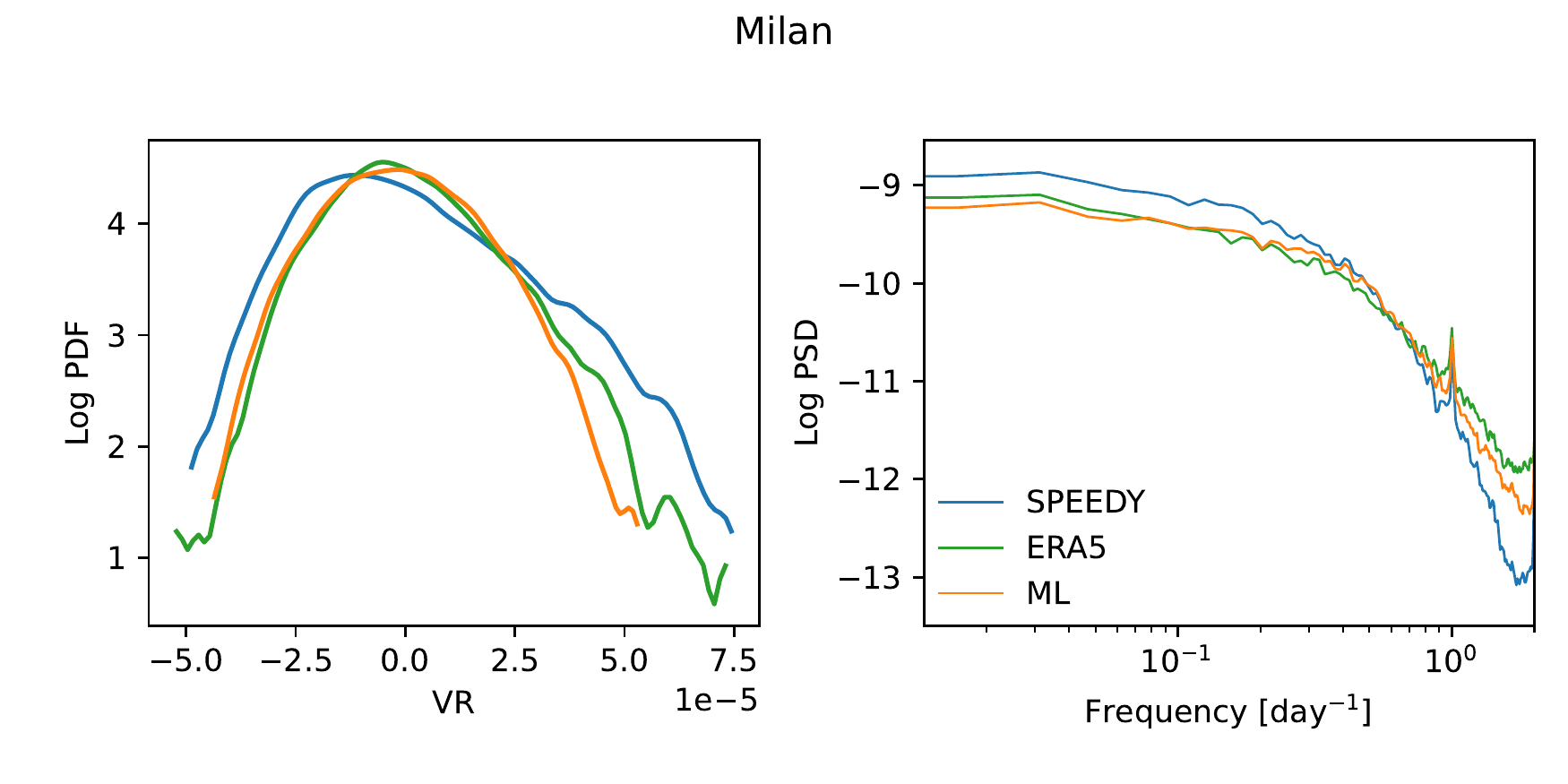}}\\
  \subfloat{\includegraphics[width=0.48\textwidth]{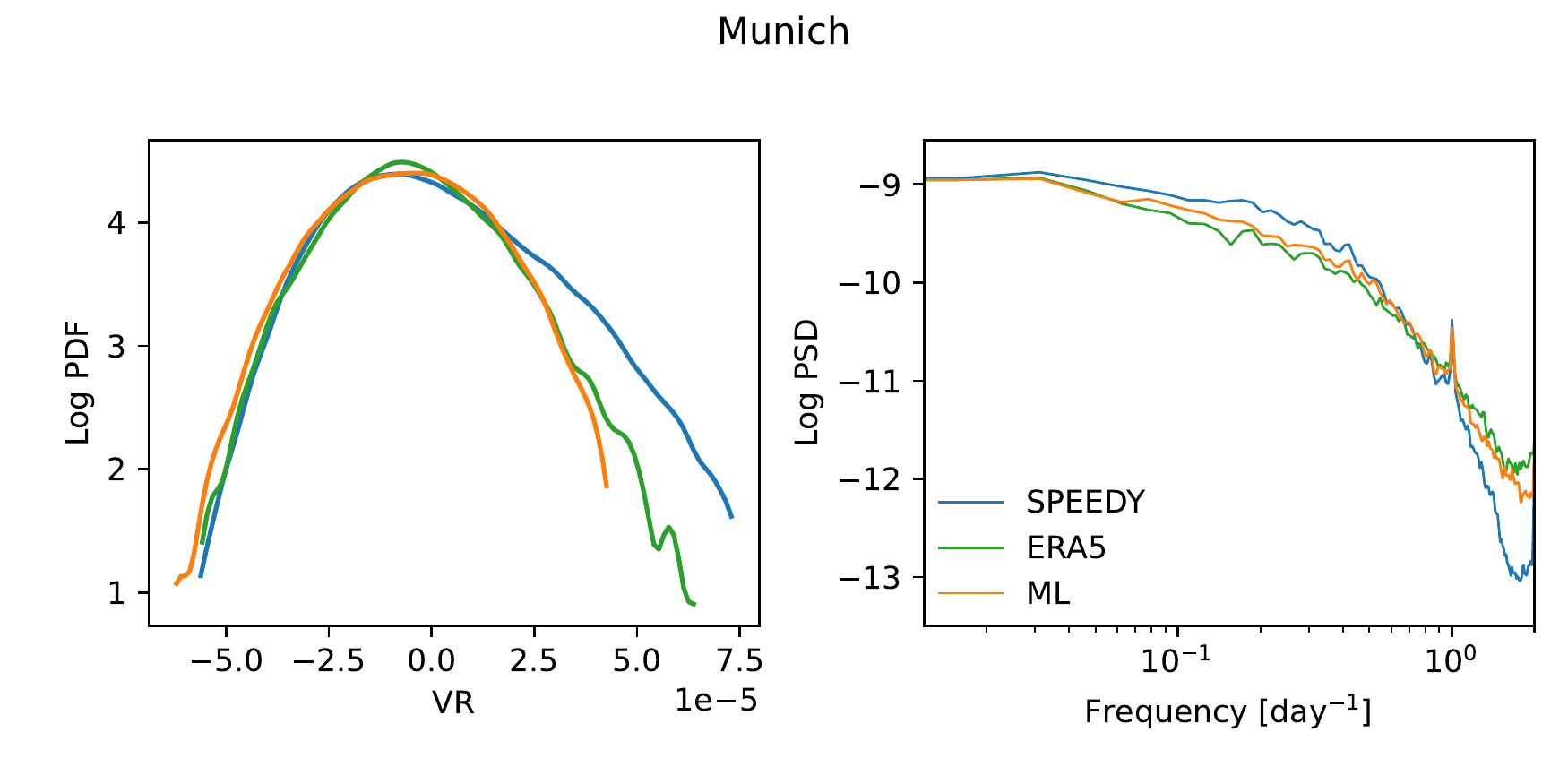}}
  \subfloat{\includegraphics[width=0.48\textwidth]{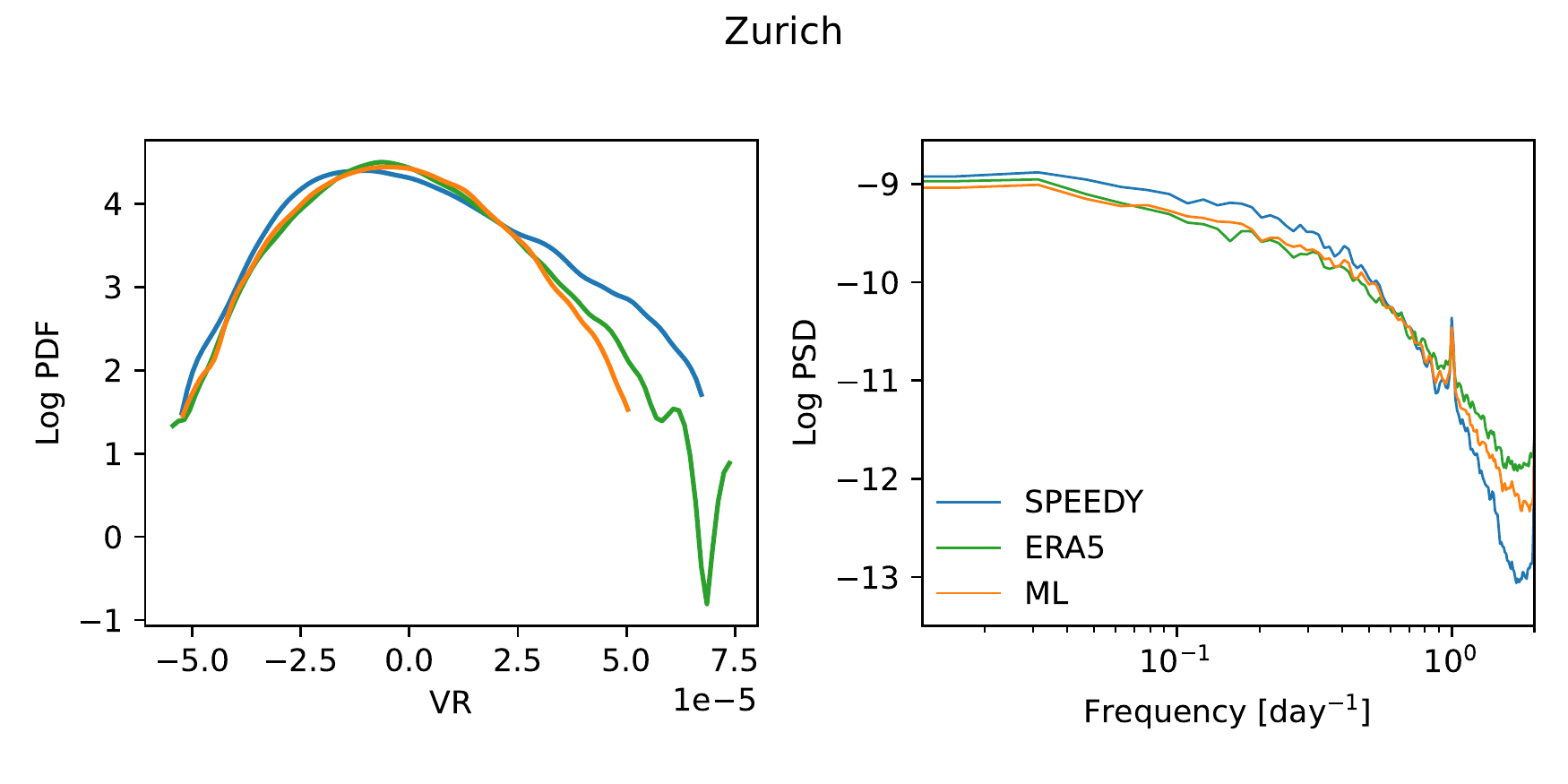}}
  \caption{Temporal statistics (left: probability density; right: spectrum amplitude) of coarse-scale vorticity at select cities computed over a 10-year period for SPEEDY GCM, machine-learning debiased simulation (ML), and reanalysis data (ERA5).}
  \label{fig:stats}
\end{figure}

\begin{figure}[t]
  \centering
  \subfloat{\includegraphics[width=0.48\textwidth]{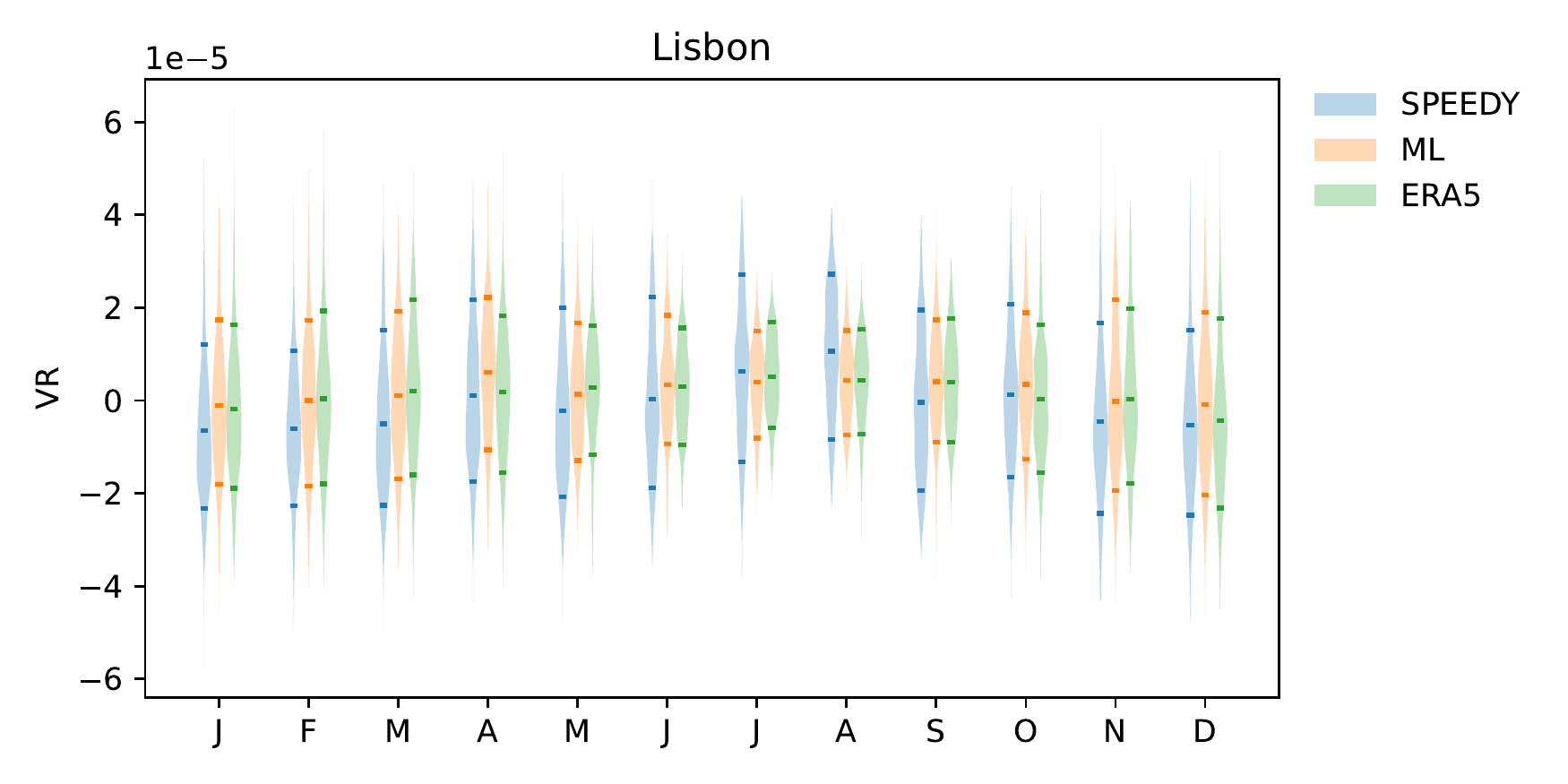}}
  \subfloat{\includegraphics[width=0.48\textwidth]{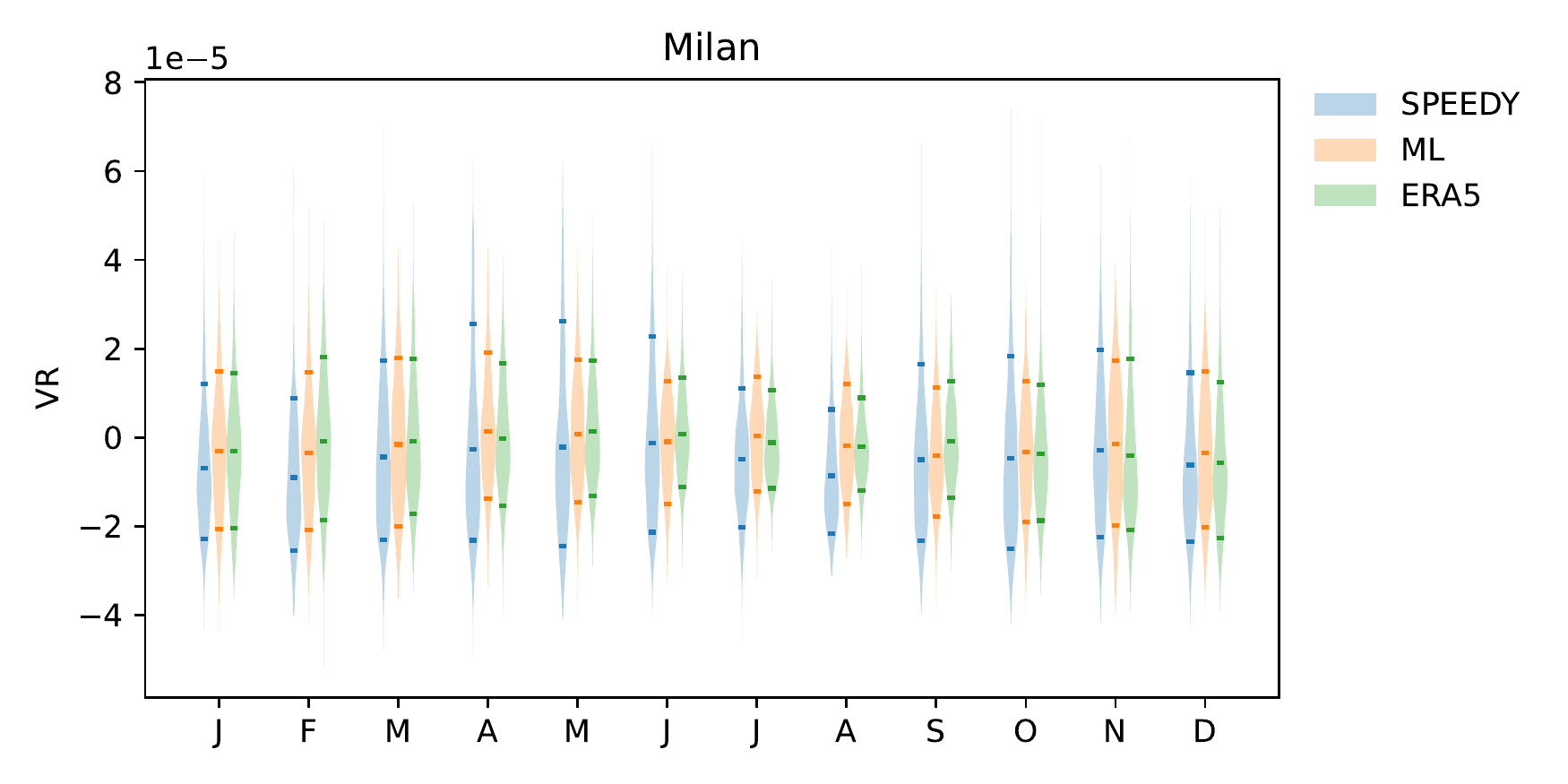}}\\
  \subfloat{\includegraphics[width=0.48\textwidth]{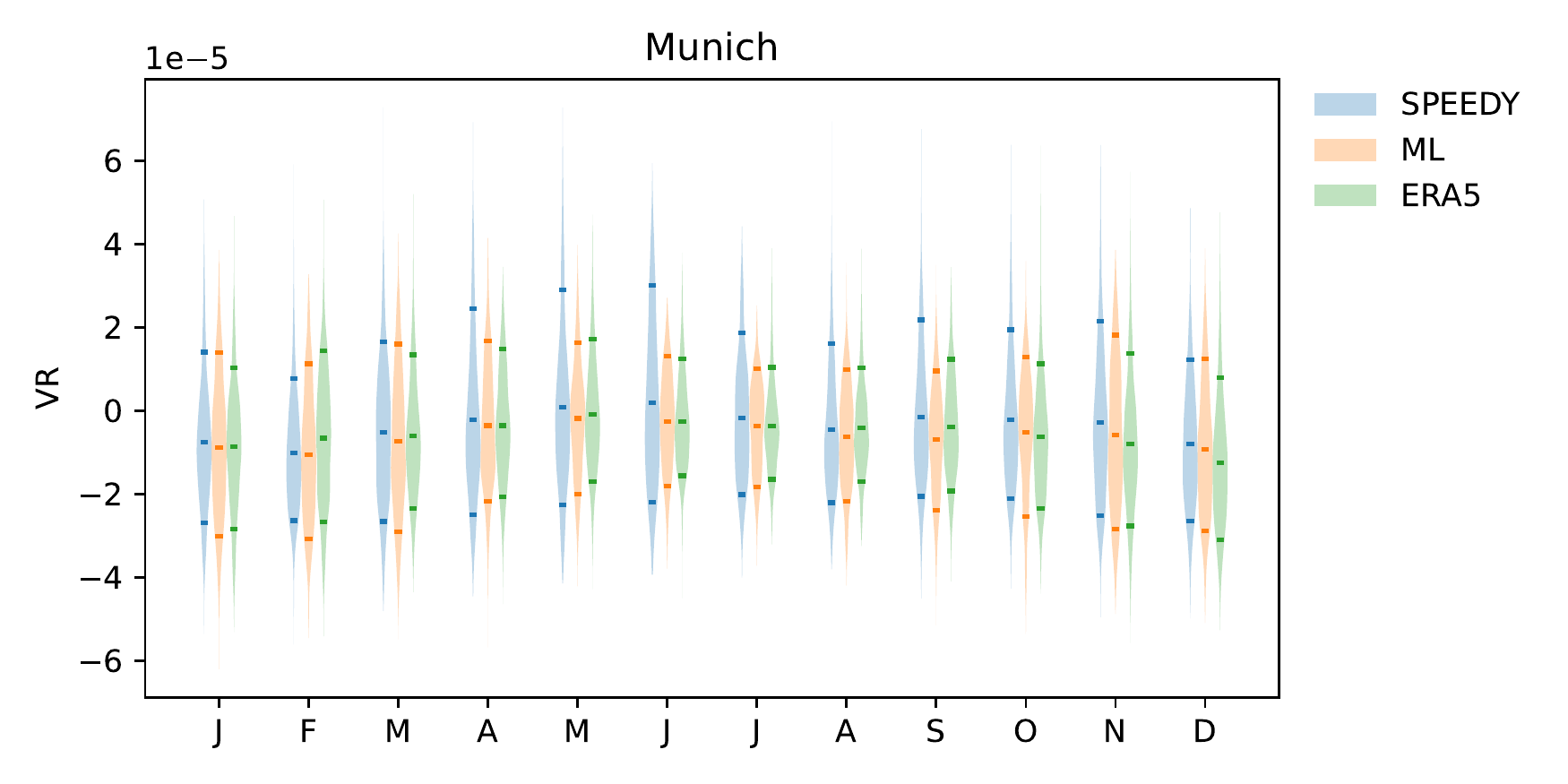}}
  \subfloat{\includegraphics[width=0.48\textwidth]{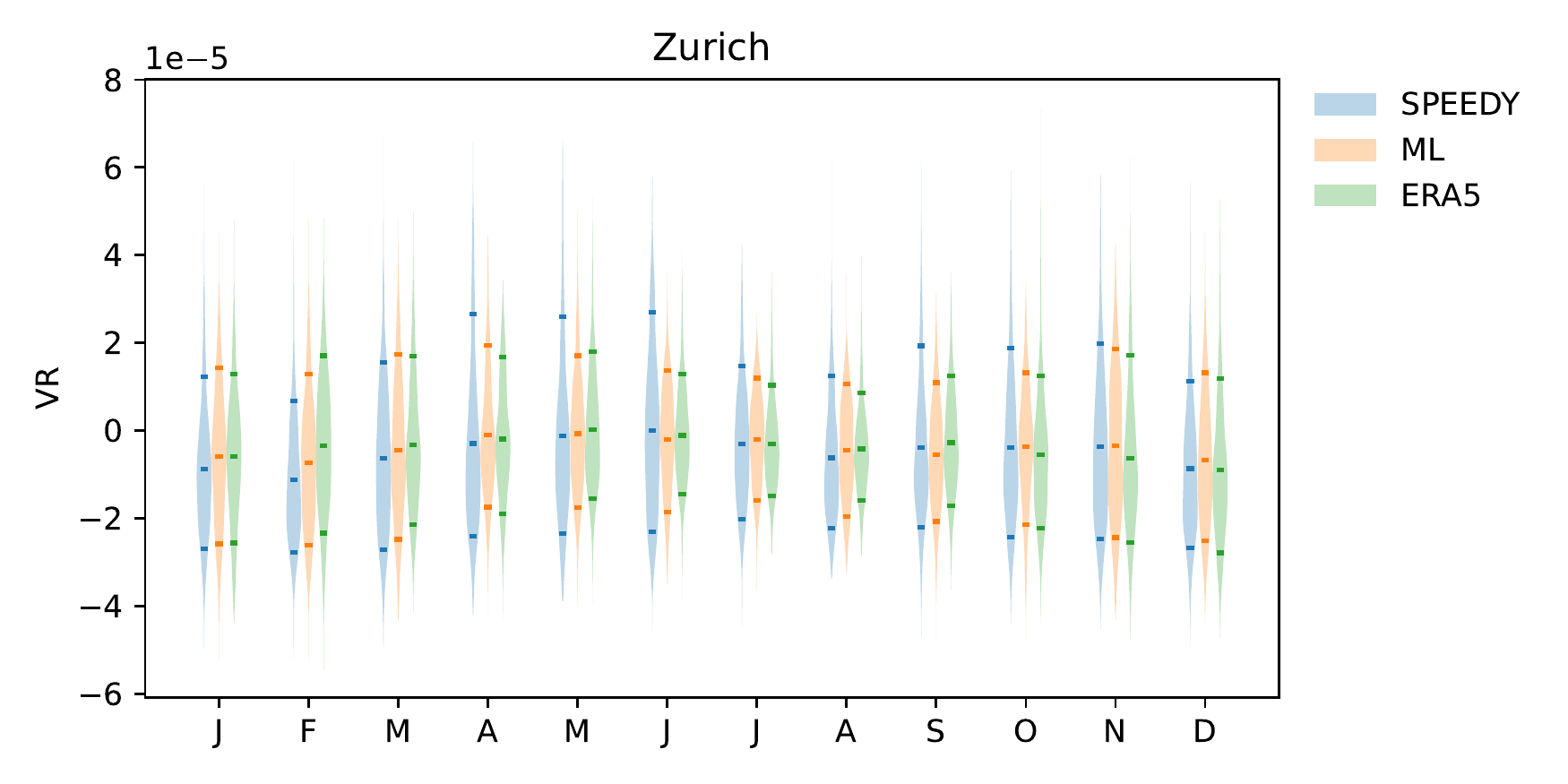}}
  \caption{Monthly distributions of coarse-scale vorticity at select cities computed over a 10-year period for SPEEDY GCM, machine-learning debiased simulation (ML), and reanalysis data (ERA5). The top, middle, and bottom notches denote the 0.9, 0.5, and 0.1 quantiles, respectively.}
  \label{fig:violins}
\end{figure}

\begin{figure}[t]
  \centering
  \includegraphics[width=0.8\textwidth]{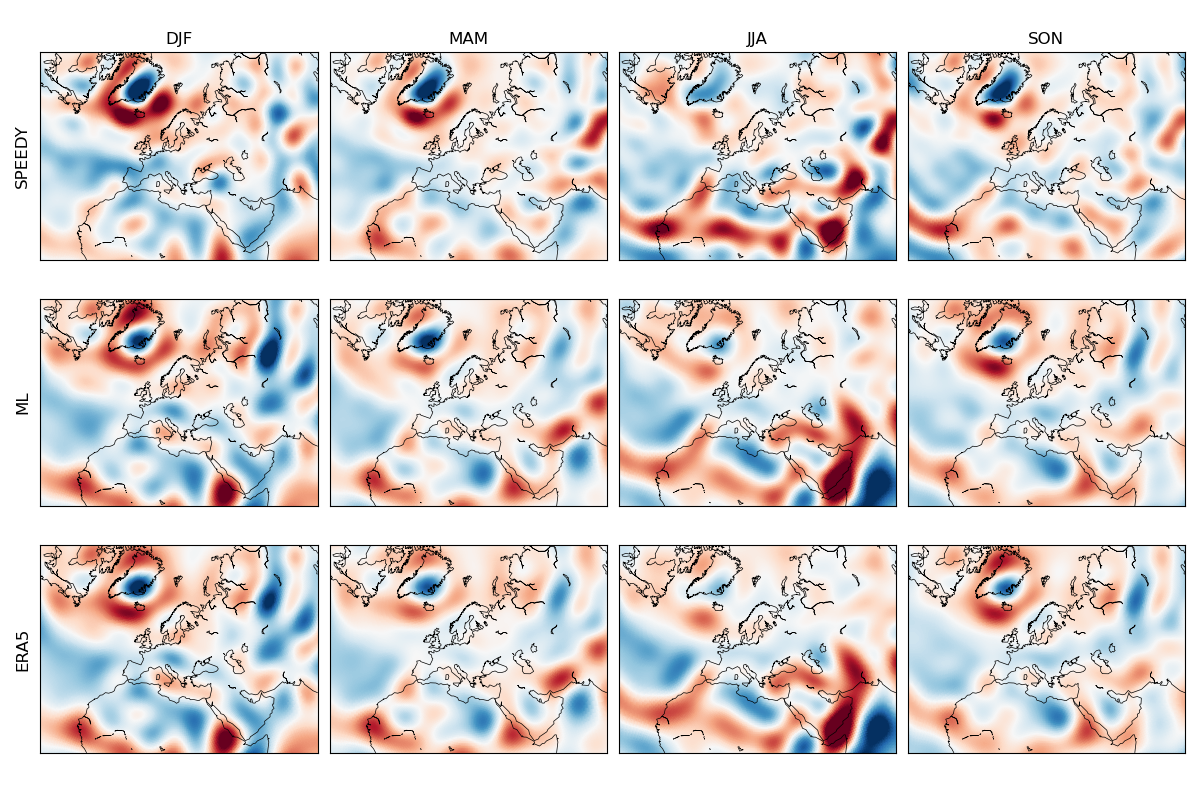}
  \caption{Seasonal averages of coarse-scale vorticity computed over a 10-year period for original SPEEDY simulation (top), machine-learning debiased simulation (middle), and reanalysis data (bottom); DJF: December--February; MAM: March--April; JJA: June--August; SON: September--November.}
  \label{fig:savg}
\end{figure}

\begin{figure}[t]
  \centering
  \subfloat[][]{\includegraphics[width=0.48\textwidth]{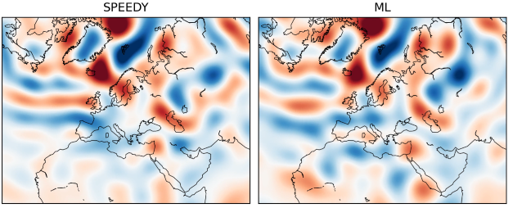}} \quad
  \subfloat[][]{\includegraphics[width=0.48\textwidth]{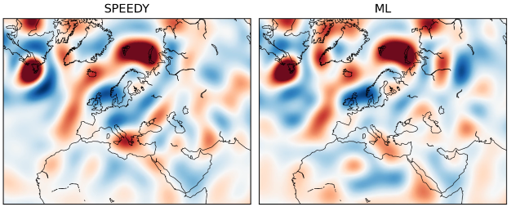}}
  \caption{Two snapshots of coarse scale vorticity for SPEEDY and the corresponding machine-learning debiased simulations (ML) resulting from Step 1 of our algorithm.  The machine learning algorithm deforms the SPEEDY simulation in order to match its statistics with observations at coarse scales but retains the overall structure of the flow.}
  \label{fig:snapp1}
\end{figure}

\paragraph{Step 1: Debiasing the coarse scales}  Figure~\ref{fig:stats} demonstrates how the debiasing algorithm is able to correct the statistical biases of the SPEEDY model in terms of temporal probability distribution function and spectrum. The agreement of the debiased output with coarse-scale observations is very good, especially in the tails of the distributions where extreme events reside.  Figure~\ref{fig:violins} shows that the machine-learning algorithm is able to correct the monthly distributions of vorticity and match them with reanalysis data.  The top and bottom monthly quantiles are captured well, which is critical for extreme events.  The monthly mean trends of the debiased output are also consistent with observations.  This is confirmed in the seasonal averages shown in Figure~\ref{fig:savg}.  Finally, in Figure~\ref{fig:snapp1} we compare two snapshots from the SPEEDY simulation with their debiased versions obtained with the machine-learning algorithm.  We clearly see that the algorithm maintains the overall consistency of the field and ``deforms'' the SPEEDY simulation as little as possible in order to correct the statistics.

\paragraph{Step 2: Reconstructing the small scales}

\begin{figure}[t]
  \centering
  \subfloat[][]{\includegraphics[width=0.48\textwidth]{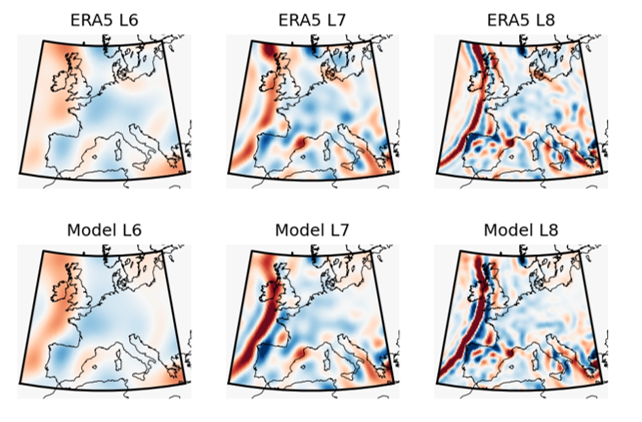}} \quad
  \subfloat[][]{\includegraphics[width=0.48\textwidth]{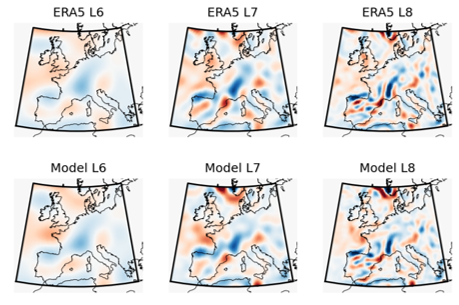}}
  \caption{Two snapshots of fine scale vorticity reconstructed from the coarse scales. The resolution-enhancing machine learning models use low resolution reanalysis data as input to show the effect of the second step in isolation.}
  \label{fig:snapp2}
\end{figure}

\begin{figure}[t]
  \centering
  \subfloat[][Probability densities]{\includegraphics[width=\textwidth]{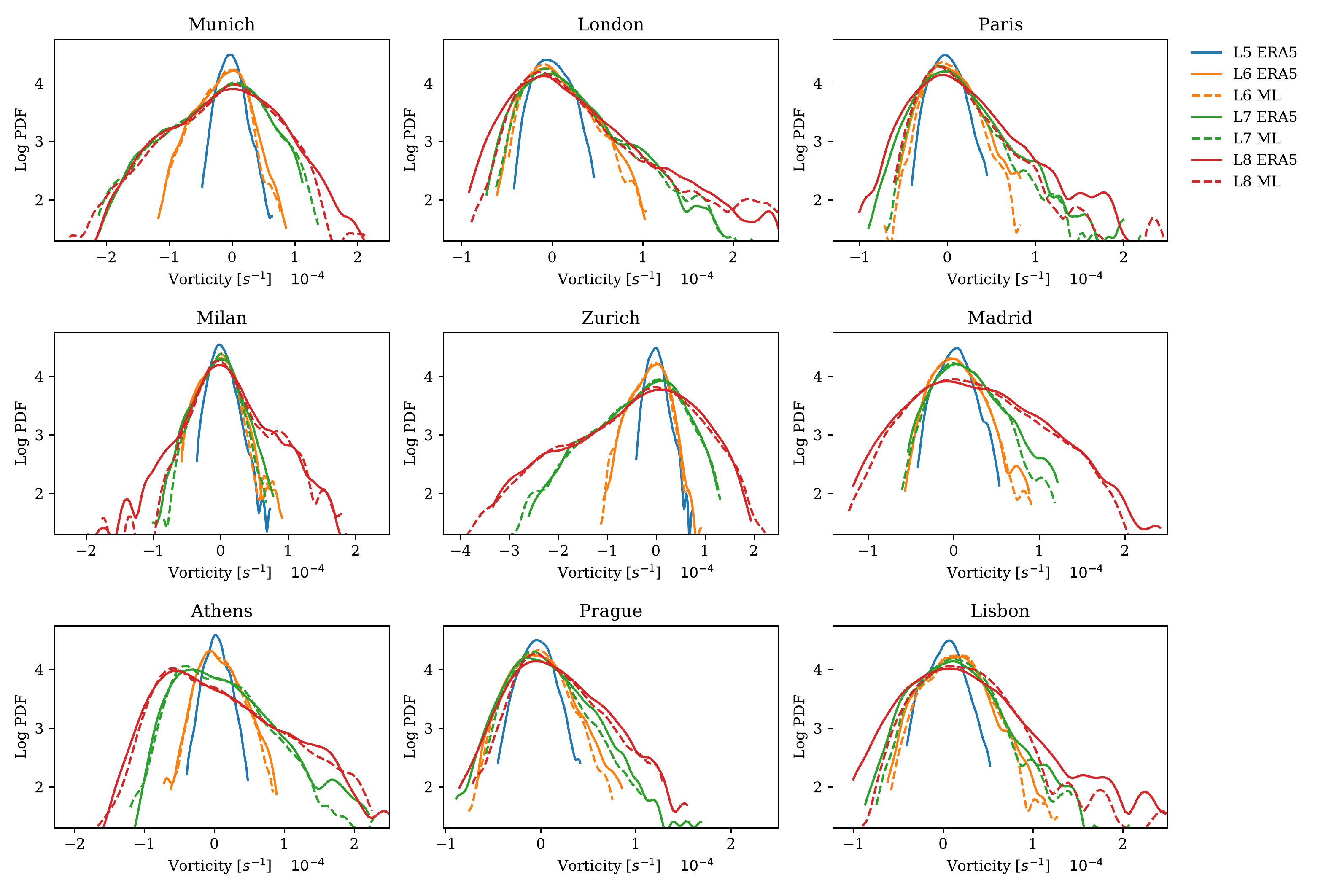}} \\
  \subfloat[][Power spectral densities]{\includegraphics[width=\textwidth]{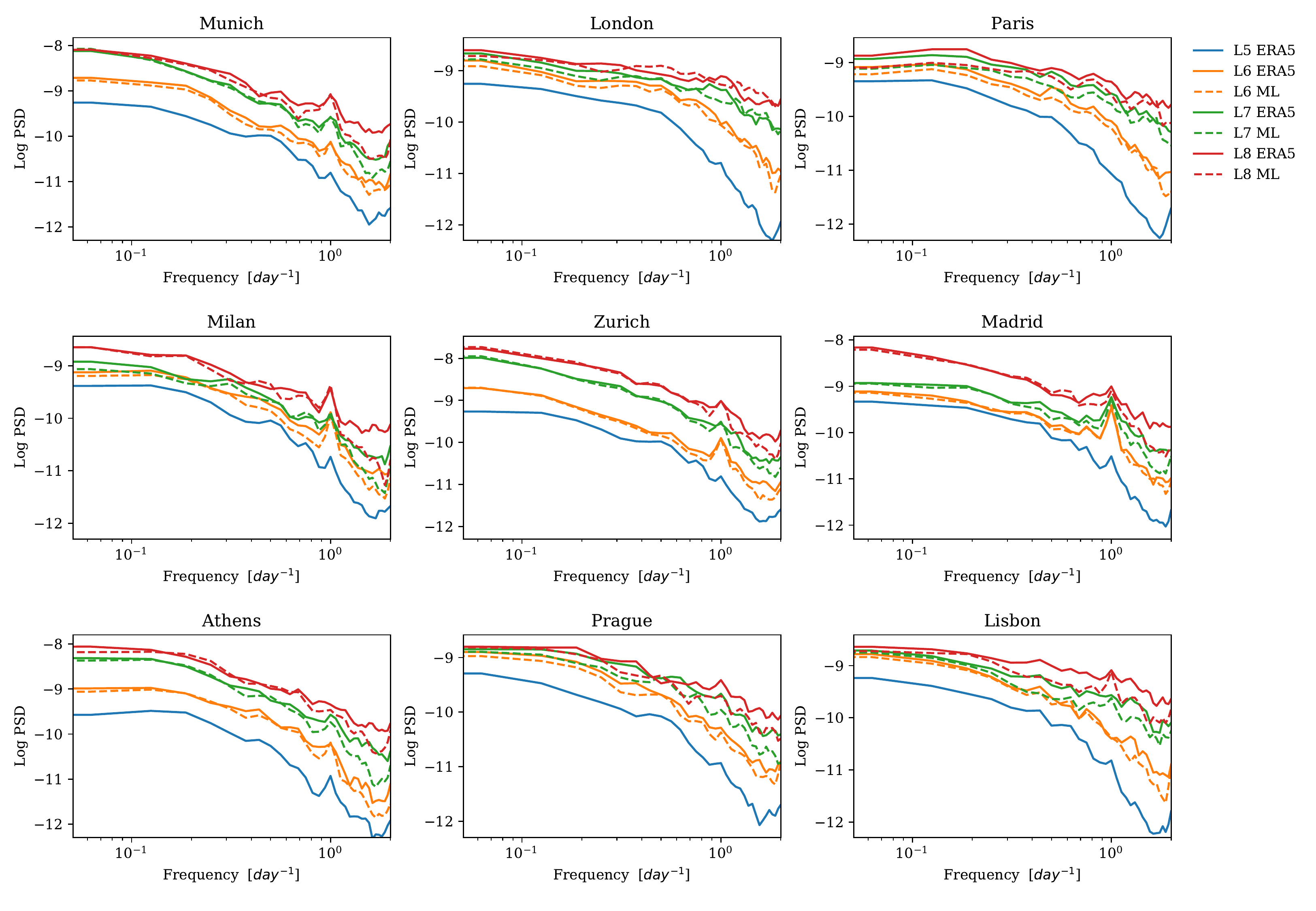}}
  \caption{Temporal statistics of vorticity computed over a one year period for select cities. The resolution-enhancing machine learning models use low resolution reanalysis data as input to show the effect of the second step in isolation.}
  \label{fig:statsera5}
\end{figure}

To test the performance of the resolution-enhancing second machine learning step in isolation, we used reanalysis data represented on the same coarse wavelet level that is used for Step 1.  Visually, Figure~\ref{fig:snapp2} shows that the algorithm is able to reconstruct the fine scales with high fidelity.  The temporal statistics of the fine scales are also captured very well; see Figure~\ref{fig:statsera5}.

\paragraph{Steps 1 and 2 combined: Generating a full-scale debiased simulation}

\begin{figure}[t]
  \centering
  \subfloat[][Probability densities]{\includegraphics[width=\textwidth]{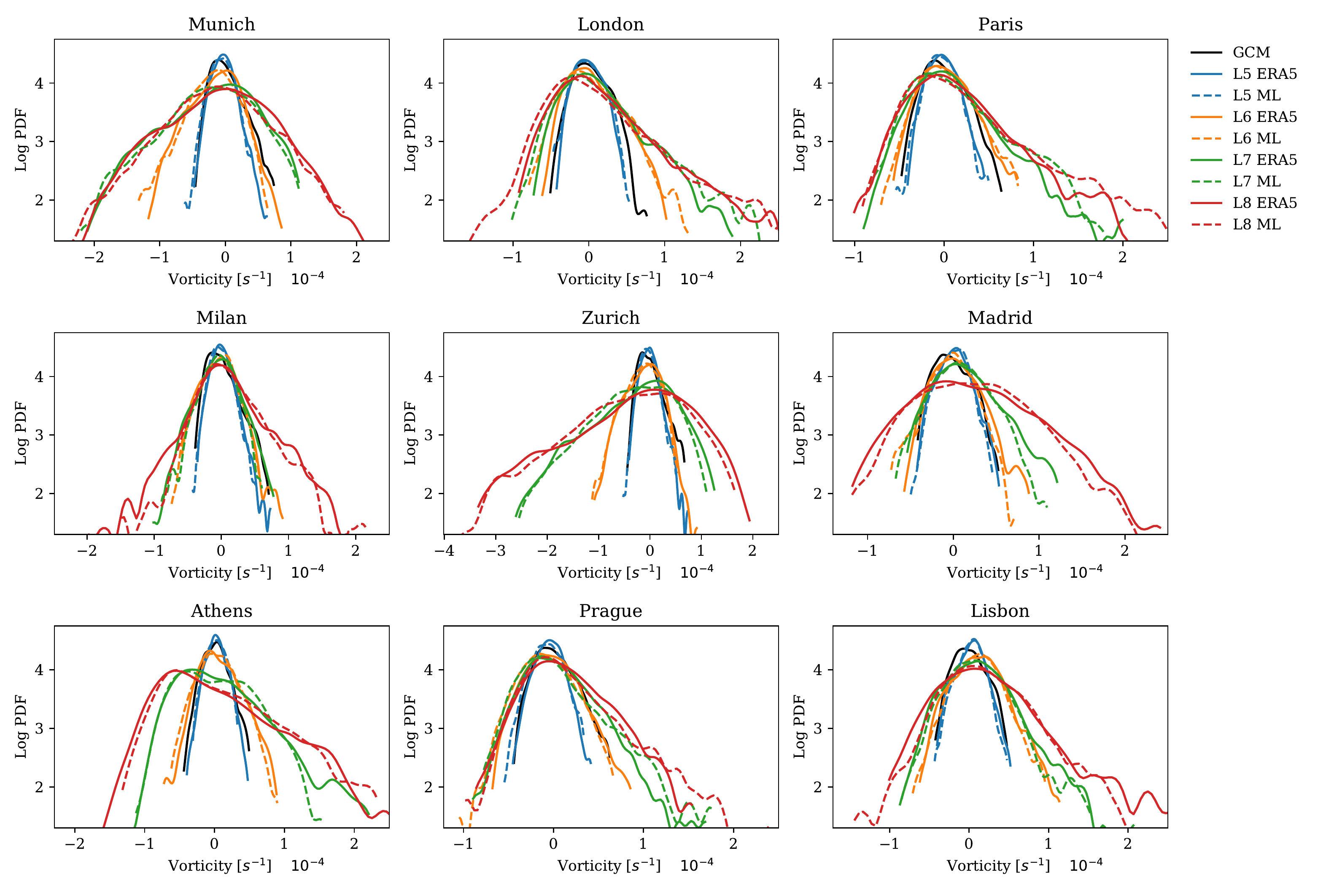}} \\
  \subfloat[][Power spectral densities]{\includegraphics[width=\textwidth]{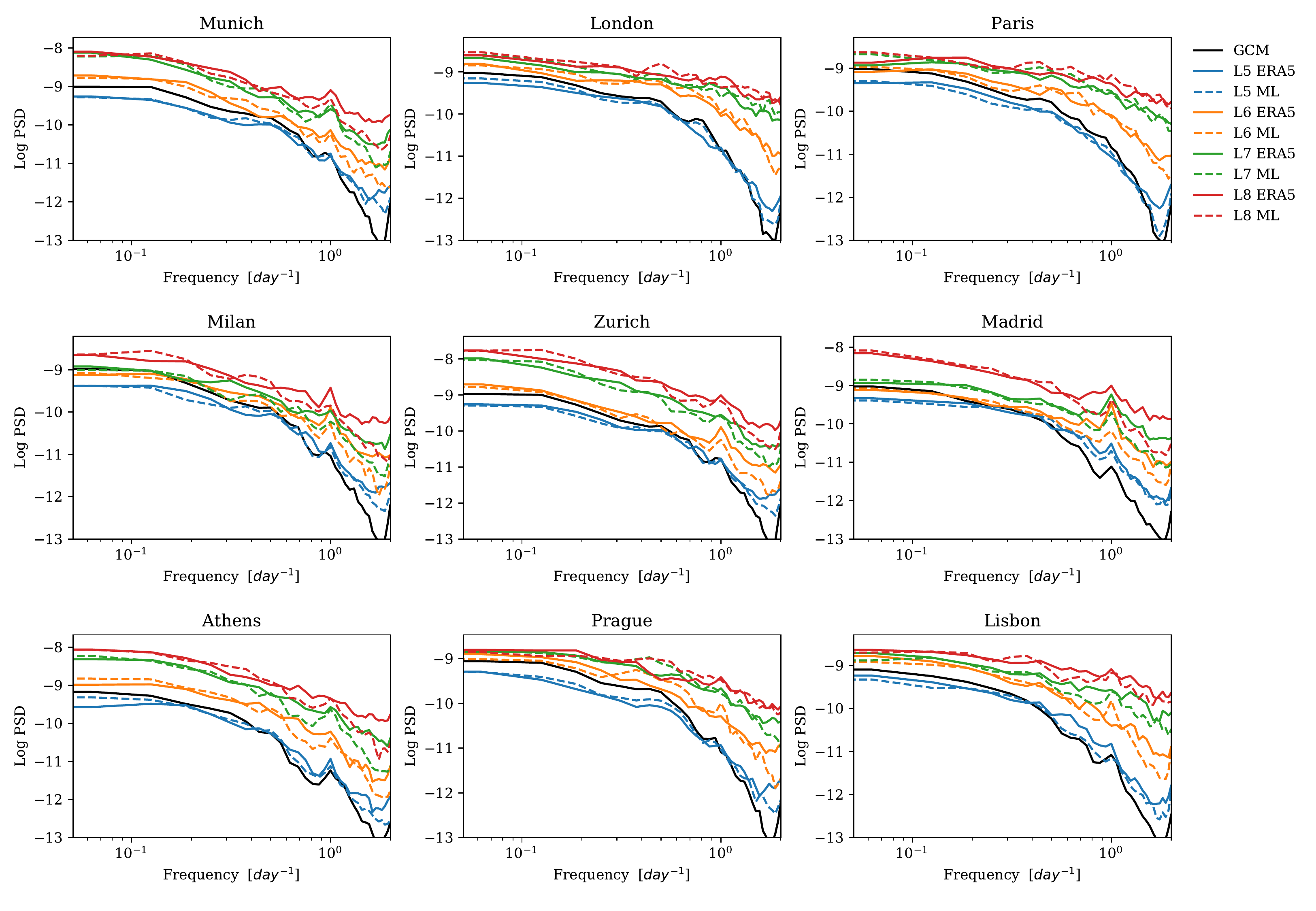}}
  \caption{Temporal statistics of vorticity computed over a one year period for select cities. he resolution-enhancing machine-learning models (``ML'') are tested on the debiased output of Step 1 to produce a full scale debiased simulation.}
  \label{fig:statsdeb}
\end{figure}

We present additional results for the full scale simulation at seven other cities besides Paris and London which are already shown in Figure~\ref{fig:results}.  The probability density functions and spectra in Figure~\ref{fig:statsdeb} show good agreement between the full resolution debiased machine learning output and observational (reanalysis) data.

\end{document}